%% file: arxiv.tex
\documentclass[10pt,twocolumn,letterpaper]{article}

\usepackage{cvpr}

\usepackage[dvipsnames,table]{xcolor}
\usepackage{times, epsfig, graphicx, amsmath, amssymb, booktabs}
\usepackage{enumitem, graphicx, multirow, amsfonts, amsmath, verbatim, algorithm, algorithmicx, algpseudocode, booktabs, array, arydshln, times, epsfig, amssymb, pifont, capt-of, wrapfig, lipsum, makecell, tabulary, pifont} 
\usepackage{etoolbox, subcaption}
\usepackage[pagebackref=true,breaklinks=true,colorlinks,bookmarks=false]{hyperref}

\usepackage[font=footnotesize,labelfont=bf,aboveskip=1pt,belowskip=-13pt]{caption}

\newdimen\abovecrulesep
\newdimen\belowcrulesep
\makeatletter
\patchcmd{\@@@cmidrule}{\aboverulesep}{\abovecrulesep}{}{}
\patchcmd{\@xcmidrule}{\belowrulesep}{\belowcrulesep}{}{}

\definecolor{demphcolor}{RGB}{144, 144, 144}
\definecolor{mygray}{gray}{0.4}
\definecolor{lightgray}{rgb}{0.9, 0.9, 0.9}
\newcommand{\demph}[1]{\textcolor{demphcolor}{#1}}

\newlength\savewidth
\newcommand\shline{\noalign{\global\savewidth\arrayrulewidth\global\arrayrulewidth 1pt}\hline\noalign{\global\arrayrulewidth\savewidth}}

\newcommand{\tablestyle}[2]{\setlength{\tabcolsep}{#1}\renewcommand{\arraystretch}{#2}\centering\footnotesize}
\makeatletter\renewcommand\paragraph{\@startsection{paragraph}{4}{\z@}{.5em\@plus1ex\@minus.2ex}{-.5em}{\normalfont\normalsize\bfseries}}
\makeatother

\newcolumntype{C}[1]{>{\centering\arraybackslash}p{#1}}
\newcolumntype{R}[1]{>{\raggedleft\arraybackslash}p{#1}}
\newcolumntype{L}[1]{>{\raggedright\arraybackslash}p{#1}}

\newcommand{\modelname}{VIOLETv2\xspace}
\newcommand{\modelorig}{VIOLET\xspace}

\newcommand{\siftgif}{68.8\xspace}
\newcommand{\sifdro}{35.4\xspace}
\newcommand{\sifdrf}{62.4\xspace}
\newcommand{\sifdrt}{74.9\xspace}
\newcommand{\sifaver}{57.6\xspace}

\usepackage{etoolbox}
\preto\align{\small}
\expandafter\preto\csname align*\endcsname{\small}
\preto\equation{\par\nobreak\small\noindent}

\def\cvprPaperID{2177}
\def\confName{CVPR}
\def\confYear{2023}

\begin{document}

\title{An Empirical Study of End-to-End \\ Video-Language Transformers with Masked Visual Modeling}
\author{Tsu-Jui Fu$^\dagger$*, Linjie Li$^\ddagger$*, Zhe Gan$^\ddagger$, Kevin Lin$^\ddagger$, William Yang Wang$^\dagger$, Lijuan Wang$^\ddagger$, Zicheng Liu$^\ddagger$\\$^\dagger$UC Santa Barbara~~$^\ddagger$Microsoft\\{\tt \small \{tsu-juifu, william\}@cs.ucsb.edu}\\{\tt \small \{lindsey.li, zhe.gan, keli, lijuanw, zliu\}@microsoft.com}}
\maketitle

\begin{abstract}
% A great challenge in video-language (VidL) modeling lies in the disconnection between fixed video representations extracted from image/video understanding models and downstream VidL data. Recent studies try to mitigate this disconnection via end-to-end training. To make it computationally feasible, prior works tend to ``imagify" video inputs, \textit{i.e.}, a handful of sparsely sampled frames are fed into a 2D CNN, followed by a simple mean-pooling or concatenation to obtain the overall video representations. Although achieving promising results, such simple approaches may lose temporal information that is essential for performing downstream VidL tasks. 
Masked visual modeling (MVM) has been recently proven effective for visual pre-training. While similar reconstructive objectives on video inputs (e.g., masked frame modeling) have been explored in video-language (VidL) pre-training,  previous studies fail to find a truly effective MVM strategy that can largely benefit the downstream performance.
% the pre-extracted video features in previous studies cannot be refined through MVM during pre-training, and thus leading to unsatisfactory downstream performance. 
In this work, we systematically examine the potential of MVM in the context of VidL learning. Specifically, we base our study on a fully end-to-end VIdeO-LanguagE Transformer (\modelorig)~\cite{fu2021violet}, where the supervision from MVM training can be backpropagated to the  video pixel space. 
In total, eight different reconstructive targets of MVM are explored, from low-level pixel values and oriented gradients to high-level depth maps, optical flow, discrete visual tokens, and latent visual features. We conduct comprehensive experiments and provide insights into the factors leading to effective MVM training, resulting in an enhanced model \modelname. Empirically, we show \modelname pre-trained with MVM objective achieves notable improvements on 13 VidL benchmarks, ranging from video question answering, video captioning, to text-to-video retrieval.\footnote{Code has been released at \url{https://github.com/tsujuifu/pytorch_empirical-mvm}.}
\end{abstract}

\section{Introduction}
%Humans are born to perceive this world from multiple modalities.
%such as vision, sound, and touch.
Video, containing multiple modalities in nature, has been used as an epitome to test how AI systems perceive. Video-language (VidL) research aims at extending this ability to convey perception via language. Popular VidL tasks were introduced, such as text-to-video retrieval~\cite{xu2016msrvtt,krishna2017dense-caption,rohrbach2015lsmdc}, video question answering~\cite{jang2017tgif-qa,xu2017msrvtt-qa}, and video captioning~\cite{xu2016msrvtt,chen2011msvd}. Recent progresses in VidL learning mostly focus on VidL pre-training~\cite{sun2019videobert,miech2019howto100m,zhu2020act-bert} with video-text matching~\cite{li2020hero,zellers2021merlot} and masked language modeling~\cite{devlin2019bert}. 
% Inspired by the advances in vision-language pre-training~\cite{chen2020uniter,tan2019lxmert},  
% Though the direct adoption of Masked Language Modeling (MLM)~\cite{devlin2019bert} has proven effective in pre-training vision-language models, 
There have also been attempts on similar masked modeling on vision inputs. For example, masked frame modeling~\cite{li2020hero} aims to recover masked frame representations. However, the pre-extracted video features cannot be refined during pre-training, which may limit its effectiveness. More recently, \modelorig~\cite{fu2021violet} designs an end-to-end video-language transformer and proposes to reconstruct discrete visual tokens for masked frame patches. Though showing some promises in recovering visual semantics, the performance improvements on downstream VidL tasks are still marginal.
% draws inspiration from masked image modeling~\cite{bao2022beit}, directly 

\begin{figure*}[t]
\centering
    \includegraphics[width=.78\linewidth]{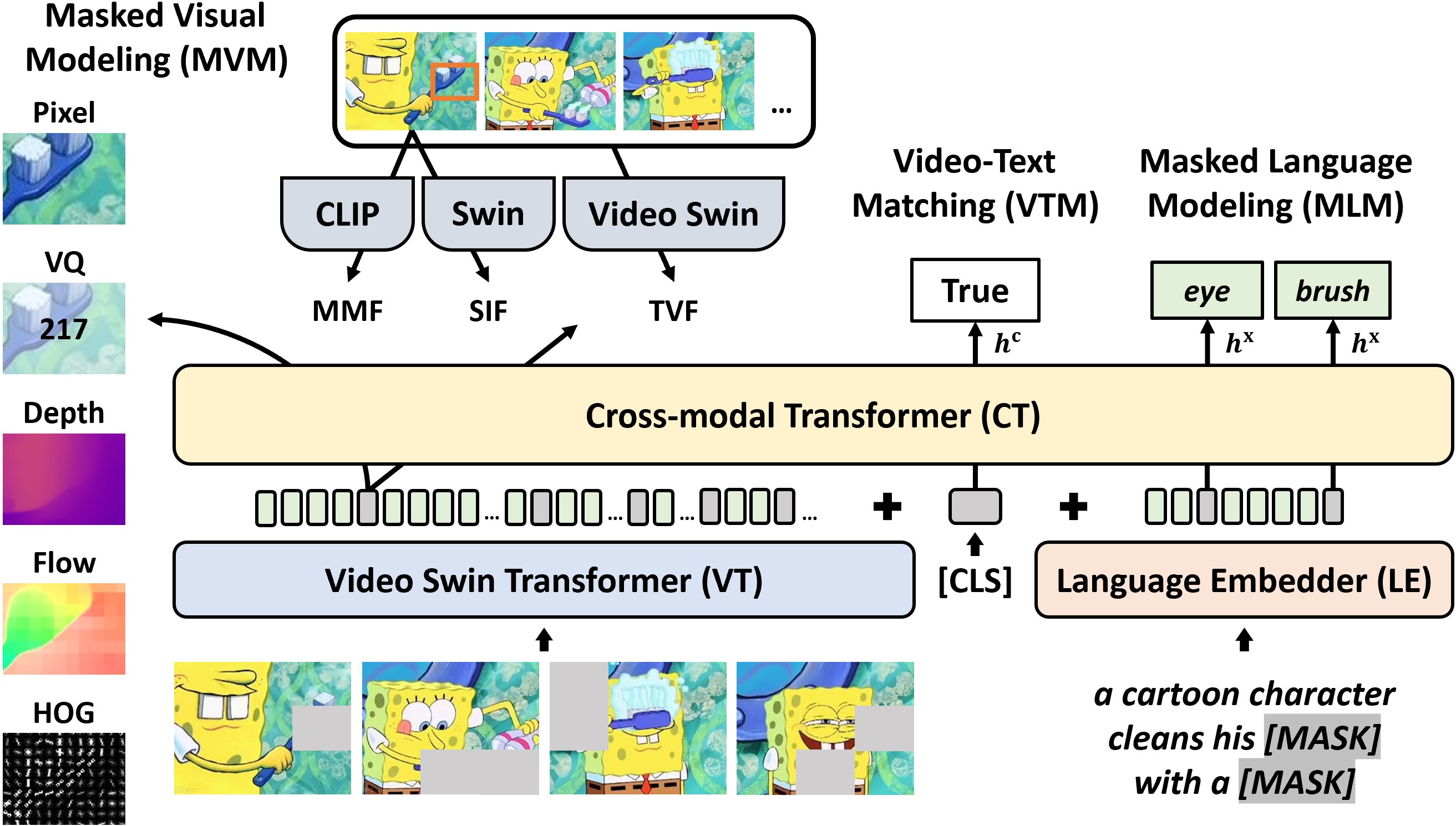}
    \vspace{0.5ex}
    \caption{We systematically explore \emph{eight} masked visual modeling (MVM) targets for end-to-end video-language (VidL) pre-training, including RGB pixel values (Pixel), histogram of oriented gradients (HOG), depth maps (Depth), optical flow (Flow), discrete visual tokens (VQ), spatial-focused image features (SIF), temporal-aware video features (TVF), and multimodal features from CLIP (MMF). Besides MVM, we pre-train VIOLET model~\cite{fu2021violet} along with video-text matching (VTM) and masked language modeling (MLM).}
    \label{fig:intro}
\end{figure*}

Meanwhile, self-supervised visual pre-training has been proven highly effective by reconstructing the masked image patches through raw pixel values~\cite{he2022mae,xie2022simmim}, discrete visual tokens~\cite{bao2022beit,zhou2022ibot}, or visual-semantic features~\cite{wei2021masked-feat,wei2022mvp}. However, they all only focus on the visual modality. It is unclear which variant of masked visual modeling (MVM) objectives can help VidL learning, especially given that the paired language inputs can already provide high-level semantics.

Motivated by this, we conduct a comprehensive study of MVM for VidL learning. As illustrated in Figure~\ref{fig:intro}, we base our study on the fully end-to-end VIdeO-LangaugeE Transformer (\modelorig)~\cite{fu2021violet}, and study a broad spectrum of MVM targets, including RGB pixel values (Pixel), histogram of oriented gradients (HOG), depth maps (Depth), optical flow (Flow), discrete visual tokens (VQ), spatial-focused image features (SIF), temporal-aware video features (TVF), and mulitmodal features (MMF). During pre-training, we mask out some proportions of the video input along both spatial and temporal dimensions, and the model learns to recover the MVM targets for these masked patches. Equipped with another two standard pre-training tasks (\emph{i.e.}, video-text matching  and masked language modeling), we empirically verify the effectiveness of different MVM variants on downstream VidL tasks.  

Our study reveals that: ($i$) spatial-focused image features (SIF) is the most  effective MVM target on video-text inputs; and ($ii$) the effects of different MVM targets on downstream VidL tasks are not shared between video-text and image-text inputs. For example, SIF extracted from the same model brings a large drop on downstream VidL performance when pre-trained with image-text pairs. In addition, we conduct comprehensive analyses of the masking strategy and ratio, combination of different MVM targets, to shed light on effective MVM training for VidL learning. We name the enhanced version of the original \modelorig~\cite{fu2021violet} with the best MVM strategy as \modelname.

Our contributions can be summarized as follows. We present an empirical study of masked visual modeling for video-language pre-training, with comprehensive analyses to reveal the ingredients for effective MVM training. \modelname with the best MVM recipe achieves strong performance on 13 VidL datasets.
% over 3 popular tasks, covering video question answering (QA), video captioning and text-to-video retrieval. 
Concretely, compared to models pre-trained on the same 5M corpus, \modelname brings mean improvements of +5.4\% accuracy on video question answering, +6.6\% recall on text-to-video retrieval, and +11.4 CIDEr on video captioning. Direct comparison to VIOLET~\cite{fu2021violet} also shows notable advantages of our model, even when pre-trained with much less data. 

% Notably, compared with VIOLET~\cite{fu2021violet} pre-trained on the same 5M pre-training dataset, \modelname brings mean improvements of +5.4\% accuracy on video question answering, +6.6\% recall on text-to-video retrieval, and +11.4 CIDEr on video captioning. 
%\linjie{update this sentence to compare with VIOLET.}
% ($i$) ($ii$)  and ($iii$) 
% , +11.4 CIDEr on video captioning,
 % and +6.6\% Recall on text-to-video retrieval.

\section{Related Work}
\noindent \textbf{Video-Language Understanding.}
Joint video-language (VidL) understanding~\cite{li2021value,liu2020collaborative-expert,jiang2020dac,le2020hcr-vqa,gabeur2020mmt,patrick2021support-set,gan2022vlp} aims at interpreting the physical world via both vision and text perception. Researchers have explored such capability on VidL tasks including text-to-video retrieval~\cite{xu2016msrvtt,krishna2017dense-caption,rohrbach2015lsmdc,lei2020tvr,li2020hero}, video question answering~\cite{jang2017tgif-qa,xu2017msrvtt-qa,lei2018tvqa,lei2020tvqa+}, moment retrieval~\cite{hendricks2017local-moment,gao2017tall,krishna2017dense-caption,lei2020tvr}, and video captioning~\cite{wang2019vatex,zhou2018youcook2,xu2016msrvtt,rohrbach2015lsmdc}. Prior arts before the large-scale pre-training era~\cite{ gao2018motion,zhang2018video-text,lei2021qvhighlights, fan2019heterogeneous,le2020hcr-vqa,lei2020tvqa+} leverage offline extracted video features~\cite{kay2017kinetics,wang2016temporal, carreira2017quo, xie2018rethinking,feichtenhofer2019slowfast,deng2009imagenet,he2016resnet,krishna2017vg,anderson2018bottom}. Later on, VidL pre-trained models~\cite{sun2019videobert,zhu2020act-bert,li2020hero,miech2019howto100m} built on the above pre-extracted features have shown promising results. To enhance the performance, there have been parallel interests in bringing in more modalities from raw video inputs~\cite{gabeur2020mmt,rouditchenko2021avlnet,liu2021hit} and end-to-end training~\cite{miech2020end,lei2021clip-bert,zellers2021merlot,bain2021frozen}, aiming to elevate video representations.
%for VidL modeling.
%till the raw pixels

\vspace{0.5ex}
\noindent \textbf{Masked Visual Modeling (MVM).}
Aligned with the success of transformer-based~\cite{vaswani2017attention} language pre-training~\cite{lan2019albert,liu2019roberta}, image-text pre-training~\cite{chen2020uniter,tan2019lxmert} and video-text pre-training~\cite{kim2021sspt-crl-vqa,yang2020bert-vqa,yang2021just-ask} have shown promising results on diverse vision-language (VL) tasks. Popular VL pre-training tasks include visual-text matching (VTM) and masked language modeling (MLM), which are directly adapted from language pre-training~\cite{devlin2019bert}. Similar masked modeling on visual inputs~\cite{chen2020uniter,li2020hero,dou2022empirical} has also been introduced to VL pre-training, but are not as useful. Among the literature of vision pre-training itself, MAE~\cite{he2022mae, tong22video-mae} and SimMIM~\cite{xie2022simmim} reconstruct the pixels of the masked image patches to enhance visual representation. BEiT~\cite{bao2022beit}, iBOT~\cite{zhou2022ibot}, VIMPAC~\cite{tan2021vimpac}, and BEVT~\cite{wang2022bevt} adopt a BERT-like pre-training strategy to recover the missing visual tokens. On the other hand, MaskFeat~\cite{wei2021masked-feat} and MVP~\cite{wei2022mvp} consider latent features for MVM, including hand-crafted HOG features and image features extracted from pre-trained CLIP models~\cite{radford2021clip}. Unlike previous studies exploring MVM on uni-modal data, in this study, we conduct a comprehensive investigation on how different MVM targets can help VidL learning.

The most relevant study to ours is VIOLET~\cite{fu2021violet}, which proposes to augment VidL pre-training with masked visual token modeling, while only showing marginal improvements on downstream performance. In contrast, our comprehensive investigation covers diverse MVM targets and studies different combinations of masking strategies, which encompasses the design of MVM as well as shows large performance improvements on downstream VidL tasks.

\input{03-method}
\input{04-experiments.tex}

\section{Conclusion}
We initiate the first empirical study on adopting masked visual modeling (MVM) for video-language (VidL) learning. We explore diverse MVM objectives upon end-to-end VIdeO-LanguagE Transformer (\modelname), including low-level pixel space, high-level visual semantics, and extracted latent features. Our results show that \modelname pre-trained on 5M video/image-text data with MVM objective achieves strong performance on 3 popular VidL tasks
% , including video question answering, video captioning and text-to-video retrieval, 
across 13 VidL benchmarks.
% We show that image-text  and video-text data may not share the same MVM target. Specifically, spatial-focused image feature (SIF) works the best on video-text inputs; only RGB pixel values can preserve the baseline performance without MVM for static images, paired with texts. 
Our comprehensive analyses on different combinations of MVM targets, various SIF target extractors, and varying masking strategies/ratios shed light on effective MVM design. We believe our study can guide future research on large-scale VidL pre-training and wish to study how MVM can generalize to larger-scale data. In addition, we vision that with the emergence of video/VidL foundation models in future works, better choices of MVM targets can be explored.
%\linjie{We believe our study can guide future research on large-scale VidL pre-training, and wish to study how MVM can generalize to larger-scale data. In addition, we vision that with the emergence of video/VidL foundation models in future works, better choices of MVM targets can be explored. 
% (similar to the use of VQ tokens/CLIP features for image-text tasks).
%}
\appendix
\input{appendix}
\clearpage
{\small
\bibliographystyle{ieee_fullname}
\bibliography{egbib}
}

\end{document}

%% file: 03-method.tex
\section{Method} \label{sec:method}
% We first introduce the problem formulation in Section~\ref{sec:prob_form}, and then describe the base model \modelorig in Section~\ref{sec:model_arch}. Finally, we discuss eight different target features considered for masked visual modeling (MVM) in Section~\ref{sec:target_feat}.

We first describe the base model \modelorig 
in Section~\ref{sec:model_arch}, and then introduce the problem formulation of our investigation in Section~\ref{sec:prob_form}. Section~\ref{sec:target_feat} discusses eight different target features for masked visual modeling (MVM).

\subsection{End-to-End Video-Language Transformer} \label{sec:model_arch}
We conduct our empirical study using an end-to-end VIdeO-LanguagE Transformer (\modelorig)~\cite{fu2021violet}, with 3 components: Video Swin Transformer (VT), Language Embedder (LE), and Cross-modal Transformer (CT). \modelorig takes video $\mathcal{V}$ and sentence $\mathcal{X}$ as inputs. Sparse-sampled frames $\{f_1, f_2, ...\}$ from $\mathcal{V}$ are first segmented into a set of video patches, and then processed by VT to compute video features $v=\{v_1, v_2, ...\}$. LE extracts the word embeddings $w=\{w_1, w_2, ...\}$ for each word token $\{x_1, x_2, ...\}$ in $\mathcal{X}$. Then, CT performs cross-modal fusion on top of $v$ and $w$ to produce joint VidL representations $h = [h^v, h^c, h^x]$, where $h^v, h^c, h^x$ denote the hidden representations of video patches, the special \texttt{[CLS]} token, and other word tokens. 

\subsection{Problem Setting} \label{sec:prob_form}
Given a large-scale video-language (VidL) dataset $D$, we aim to pre-train a VidL transformer to learn effective video-text representations. The learned representations can be transferred to downstream tasks for performance improvement. Different from existing works that focus on MVM for pure vision problems~\cite{bao2022beit,he2022mae,zhou2022ibot}, we study MVM as a VidL pre-training task. Given a video-text pair $(\mathcal{V},\mathcal{X})$ where $\mathcal{V}$ is a sequence of video frames and $\mathcal{X}$ is a sequence of word tokens. As shown in Figure~\ref{fig:intro}, we randomly mask out some portions of the input frames $\mathcal{V}$, and learn to predict the target features corresponding to the masked patches. To output a correct prediction, the model will have to resort to other relevant video frames $\mathcal{V}$ and/or text tokens $\mathcal{X}$. This facilitates cross-modality learning for better VidL understanding.

In addition, we employ the commonly used VidL pre-training objectives, including video-text matching (VTM) and masked language modeling (MLM), where VTM aims to predict whether an input video-text pair is matched or not, while MLM aims to predict the masked word tokens from the surrounding context.\footnote{Refer to the Appendix for detailed formulation of VTM and MLM.} Our overall pre-training objective can be written as:
\begin{equation}
\begin{aligned} \label{eqn:all-loss}
    \mathcal{L} = \mathcal{L}_\text{MVM} + \mathcal{L}_\text{VTM} + \mathcal{L}_\text{MLM},
\end{aligned}
\end{equation}
where $\mathcal{L}_\text{MVM}$, $\mathcal{L}_\text{VTM}$, $\mathcal{L}_\text{MLM}$ are the MVM, VTM and MLM objectives, respectively.

\subsection{Target Features} \label{sec:target_feat}
Masked visual modeling (MVM) is a generic masked feature prediction task, where we mask out some of the visual input patches, and then predict the target features corresponding to the masked ones. Thus, a core design of MVM is the target features, which enables \modelorig learning a desired aspect of visual modeling. % a desired prediction capability. 
While MVM has been explored in pure vision tasks~\cite{bao2022beit,he2022mae,wei2021masked-feat}, it remains an open question whether MVM can facilitate the interactions between video and language modalities. In this study, we investigate \textit{what design of MVM is effective in the context of video-language pre-training?} 

Following~\cite{xie2022simmim,wei2021masked-feat}, we employ a simple linear layer or 2-layer MLP as the prediction head for MVM, to project the hidden video representations ($h^v$, of hidden size 768) from CT to the same dimension as the MVM targets. The default MVM loss is the $l_1$ loss, unless specified otherwise. Next, we introduce the considered target features in details.

\vspace{0.5ex}
\noindent \textbf{RGB Pixel Values (Pixel).} We treat the normalized RGB pixel values as a candidate target feature. During MVM, \modelorig learns to reconstruct the pixel values of the masked patches. The linear MVM head projects $h^v$ into the same dimension as the raw video frame patch ($ H\times W \times 3$).

\vspace{0.5ex}
\noindent \textbf{Histogram of Oriented Gradients (HOG).} HOG~\cite{dalal2005hog} is a pioneer feature descriptor that describes the gradients of orientations of the image. 
While HOG has been proven effective for visual pre-training~\cite{wei2021masked-feat}, it is unknown whether it can benefit VidL pre-training. We extract HOG features in a dense grid level, and use such feature descriptors as the prediction targets of MVM. The HOG feature map is of the same size as the input video frame, but with channel size 1. The linear MVM prediction head projects $h^v$ to the same dimension as HOG for the video frame patch ($ H\times W \times 1$).

\vspace{0.5ex}
\noindent \textbf{Depth Maps (Depth).} Since depth maps usually contains finer-grained details of the object shapes and general scene layout of the foreground objects, it is worth exploring whether depth maps can be used to improve the scene/object understanding capability of a VidL pre-trained model. To obtain such MVM target, we employ a pre-trained dense prediction transformer (DPT)~\cite{ranftl2021dpt} to perform monocular depth estimation given an input video frame. The linear prediction head used for Depth is the same as the one for HOG, as both targets are of channel size 1.
\input{tabs/mvm_target_on_webvid.tex}

\vspace{0.5ex}
\noindent \textbf{Optical Flow (Flow).} Optical flow is commonly used in motion analysis and video understanding. Here, we analyze whether apparent velocity of objects can benefit VidL pre-training. We employ a pre-trained recurrent all-pairs field transforms (RAFT)~\cite{teed2020raft} to compute optical flow given the consecutive video frames. We directly use the estimated optical flow values as the prediction target, and supervise the MVM training with $l_1$ loss.  To obtain the MVM predictions, we concatenate the hidden video representations computed by CT on consecutive frames, and employ a linear layer to project the concatenated video representations (of hidden size 768 $\times$ 2) to the same dimension as the estimated optical flow target for a given patch ($H\times W \times 2$).

\vspace{0.5ex}
\noindent \textbf{Discrete Visual Tokens (VQ).} In addition to continuous MVM targets, we also consider the discrete variational autoencoder (dVAE)~\cite{oord2017vq-vae,ramesh2021dalle} to quantize video inputs. dVAE is learned to tokenize images into discrete visual tokens $q$ from a finite dictionary, and then reconstruct the original visual scene based on $q$, where $q$ should have a one-to-one correspondence with the input image patches spatially. We first adopt dVAE to tokenize the $t^\text{th}$ video frame $f_t$ into $q_t$: $q_t = \text{dVAE}(f_t)$, and then a 2-layer MLP is used to project $h_v$ into the finite VQ vocabularies. As VQ token is discrete, we can model MVM with VQ as a classification problem, and adopt the cross-entropy loss to optimize the MVM training, following~\cite{bao2022beit,fu2021violet}. 

\vspace{0.5ex}
\noindent \textbf{Spatial-focused Image Features (SIF).} We investigate whether image features can be useful for improving VidL pre-training. We employ a well-known vision transformer (such as Swin Transformer~\cite{liu2021swin}) to extract the grid features given an input image. We then normalize the extracted grid features and consider them as ground-truth MVM targets. Likewise, we adopt a 2-layer MLP to project $h_v$ to the same dimension as the image feature target.

\vspace{0.5ex}
\noindent \textbf{Temporal-aware Video Features (TVF).} We also study the impact of video features to VidL pre-training. We employ pre-trained video transformer (such as Video Swin Transformer~\cite{liu2022video-swin}) to compute temporal-aware features for this analysis.  
Given a set of video frames, we use the transformer to extract video features in the form of space-time cubes, and then apply $l_1$ regression between normalized video features and MVM predictions from a 2-layer MLP head of the masked video patches.

\vspace{0.5ex}
\noindent \textbf{Multimodal Features (MMF).} We further study if the features learned via multimodal pre-training can benefit VidL pre-training. We utilize the vision branch of the ViT-Base backbone~\cite{dosovitskiy2021vit} in CLIP~\cite{radford2021clip} to extract such multimodal features, and use the normalized features as the prediction targets in MVM pre-training. Again, we apply $l_1$ regression between the MVM predictions made via a 2-layer MLP head and the MMF targets for the masked patches.

In the following sections, we conduct comprehensive investigation over MVM targets described above, and perform detailed analysis on MVM strategies. To avoid confusion, we denote the strongest model with the most effective MVM training as \modelname.

%% file: tabs/mvm_target_on_webvid.tex
% \begin{table}[t]
% \centering
%     \tablestyle{3pt}{1.1} 
%     \def \w{20pt} 
%     % \resizebox{\linewidth}{!}{
%     \begin{tabular}{ll| ccccc}
%         \shline
%         Pre-training & \multirow{2}{*}{MVM Target} & TGIF-Frame & \multicolumn{4}{c}{DiDeMo-Retrieval} \\
%         \cmidrule(lr){3-3}\cmidrule(lr){4-7}
%         Task &  & Acc. & R1 & R5 & R10 & AveR \\
%         \hline
%         % None & None &  \\
%         VTM+MLM & None &  68.1 &  28.7 & 57.0 & 69.7 & 51.8\\
%         \hline
%         \multirow{8}{*}{+MVM} & Pixel & 68.3 & 29.2 & 58.6 & 70.1 & 52.6 \\
%           & HOG~\cite{dalal2005hog} & 67.3 & 26.6 & 54.9 & 68.1 & 49.8\\
%          \cline{2-7}
%          & VQ Token (DALL-E~\cite{ramesh2021dalle}) & 68.4 & 28.1 & 56.6 & 69.4 & 51.3  \\
%           & Multimodal tokens (CLIP~\cite{radford2021clip}) & 67.7 & 29.8 & 57.8 & 68.5 & 52.1\\
%          \cline{2-7} 
%         & Depth (DPT~\cite{Ranftl2021dpt}) & 68.0 & 27.3 & 55.0 & 68.3 & 50.2 \\
%          & Optical Flow (RAFT~\cite{v}) & 67.6 & 30.3 & 58.0 & 70.3 & 52.9\\
%           & 2D Feature (Swin~\cite{liu2021swin}) & \textbf{68.8} & \textbf{35.1} & \textbf{63.3} & \textbf{73.1} & \textbf{57.2} \\
%           & 3D Feature (VidSwin~\cite{liu2022ideo-swin}) & 67.5 & 25.8 & 55.0 & 68.0 & 49.6  \\
%         \shline
%     \end{tabular}
%     % }
%     \caption{\textbf{Comparing target features for MVM (video)}. All
% variants are pre-trained on WebVid~\cite{bain2021frozen} for 5 epochs.}
%     \vspace{-2ex}
%     \label{table:mvm-webvid}
% \end{table}

\begin{table*}[t]
\centering
    \tablestyle{6pt}{1.0} 
    \def \w{20pt} 
    % \resizebox{\linewidth}{!}{
    \begin{tabular}{ll| ccccc}
        \shline
        \multirow{2}{*}{Pre-training Tasks} & \multirow{2}{*}{MVM Target} & TGIF-Frame & \multicolumn{4}{c}{DiDeMo-Retrieval} \\
        \cmidrule(lr){3-3}\cmidrule(lr){4-7}
         &  & Acc. & R1 & R5 & R10 & AveR \\
        \hline
        % None & None &  \\
        VTM+MLM & None &  68.1  &  28.7 & 57.0 & 69.7 & 51.8\\
        \hline
        \multirow{7}{*}{+MVM} & RGB Pixel Values & 68.3 {\color{ForestGreen}(+0.2)} & 29.2 {\color{ForestGreen}(+0.5)} & 58.6 {\color{ForestGreen}(+1.6)} & 70.1 {\color{ForestGreen}(+0.4)}& 52.6 {\color{ForestGreen}(+0.8)}\\
          & Histogram of Oriented Gradients~\cite{dalal2005hog} & 67.3 {\color{BrickRed}(-0.8)} & 26.6 {\color{BrickRed}(-2.1)} & 54.9 {\color{BrickRed}(-2.1)} & 68.1 {\color{BrickRed}(-1.6)}& 49.8 {\color{BrickRed}(-2.0)}\\
          \cline{2-7} 
        & Depth Maps (DPT-L~\cite{ranftl2021dpt}) & 68.0 {\color{BrickRed}(-0.1)} & 27.3 {\color{BrickRed}(-1.4)} & 55.0 {\color{BrickRed}(-2.0)}& 68.3 {\color{BrickRed}(-1.4)}& 50.2 {\color{BrickRed}(-1.6)}\\
         & Optical Flow (RAFT-L~\cite{teed2020raft}) & 67.6 {\color{BrickRed}(-0.5)} & 30.3 {\color{ForestGreen}(+1.6)} & 58.0 {\color{ForestGreen}(+1.0)}& 70.3 {\color{ForestGreen}(+0.6)}& 52.9 {\color{ForestGreen}(+1.1)}\\
         \cline{2-7} 
         \rowcolor{lightgray}
          & Spatial-focused Image Features (Swin-B~\cite{liu2021swin}) & \textbf{\siftgif \space {\color{ForestGreen}(+0.7)}} & \textbf{ \sifdro \space {\color{ForestGreen}(+6.7)}} & \textbf{\sifdrf \space {\color{ForestGreen} (+5.4)}} & \textbf{\sifdrt \space {\color{ForestGreen}(+5.2)}} & \textbf{\sifaver \space {\color{ForestGreen}(+5.8)}} \\
          % & Temporal-aware Video Features (VidSwin-B~\cite{liu2022ideo-swin}) & 67.5 {\color{BrickRed}(-0.6)} & 25.8 {\color{BrickRed}(-2.9)} & 55.0 {\color{BrickRed}(-2.0)} & 68.0 {\color{BrickRed}(-1.7)} & 49.6 {\color{BrickRed}(-2.2)} \\
          & Temporal-aware Video Features (VidSwin-L~\cite{liu2022video-swin}) & 68.0 {\color{BrickRed}(-0.1)} &  32.8 {\color{ForestGreen}(+4.1)} &  60.5 {\color{ForestGreen}(+3.5)} &  73.0 {\color{ForestGreen}(+3.3)} &  55.4 {\color{ForestGreen}(+3.6)} \\
          % {'r@1': '32.79', 'r@5': '60.47', 'r@10': '73.03', 'median': '3'}
          \cline{2-7}
         & Discrete Visual Tokens (DALL-E~\cite{ramesh2021dalle}) & 68.4 {\color{ForestGreen}(+0.3)} & 28.1 {\color{BrickRed}(-0.6)} & 56.6 {\color{BrickRed}(-0.4)} & 69.4 {\color{BrickRed}(-0.3)} & 51.3 {\color{BrickRed}(-0.5)} \\
          & Multimodal Features (CLIP-ViT-B~\cite{radford2021clip}) & 67.7 {\color{BrickRed}(-0.4)} & 29.8 {\color{ForestGreen}(+1.1)} & 57.8 {\color{ForestGreen}(+0.8)}& 68.5 {\color{BrickRed}(-1.2)} & 52.1 {\color{ForestGreen}(+0.3)}\\
        \shline
    \end{tabular}
    % }

    \caption{\textbf{Comparing target features for MVM applied to video-text data}. All variants are pre-trained on WebVid~\cite{bain2021frozen} for 5 epochs. Masking is performed randomly (RM) with ratio of 15\%. The final pre-training setting is highlighted in \colorbox{lightgray}{gray}. }
    \label{table:mvm-webvid}
\end{table*}

%% file: 04-experiments.tex
\section{Study: Target Features for MVM} \label{sec:study}
\noindent \textbf{Settings.} We conduct pre-training on WebVid-2.5M~\cite{bain2021frozen} for 5 epochs, and report accuracy on TGIF-Frame~\cite{jang2017tgif-qa} for video question answering and R1/R5/R10/AveR on DiDeMo~\cite{hendricks2017didemo} for text-to-video retrieval.\footnote{We base our ablation experiments on these two representative datasets for fast iteration, our main results are reported on 13 benchmarks in Section~\ref{sec:sota}. Details about downstream adaptation are included in the Appendix.} We initialize our Video Swin Transformer (VT) with VideoSwin-Base~\cite{liu2022video-swin}, pre-trained on Kinetics-600~\cite{kay2017kinetics}. Language Embedder (LE) and Cross-modal Transformer (CT) are initialized from pre-trained BERT-Base~\cite{devlin2019bert}. During pre-training, we sparsely sample  4 video frames and randomly crop them into 224x224 to split into patches with $H$ = $W$ = 32. For all downstream tasks, we adopt the same video frame size and patch size but 5 sparse-sampled frames. We keep the training recipe (\textit{e.g.}, optimizer settings, masking ratio, training schedule, \emph{etc.}) consistent across all targets, which we find generally good in practice.\footnote{Refer to the Appendix for more on training details.} For MVM targets that involve a teacher model, we use official models released by the authors. We compare models pre-trained with 8 different MVM variants to the baseline pre-trained with only VTM and MLM. Our goal is to find the best MVM target features that can provide the largest performance improvement over this baseline. Results are summarized in Table~\ref{table:mvm-webvid}. We first categorize the MVM targets into 4 groups, and discuss their performance in details.

\input{tabs/combine_target.tex}

\vspace{0.5ex}
\noindent \textbf{One-stage Visual Targets.} We include \textit{Pixel} and \textit{HOG}, as they do not require training a deep neural network in advance to extract these features. Compared to the baseline without MVM objective, regressing the explicit RGB colors contributes to a relatively small gain of +0.2\% on TGIF-Frame and +0.8\% on AveR for DiDeMo Retrieval. In contrast, HOG renders degradation on downstream video-language (VidL) performance (-0.8\% on TGIF-Frame and -2.0\% on DiDeMo-Retrieval). We hypothesize that this is due to the missing color information in HOG features, which is critical in VidL understanding. 

\vspace{0.5ex}
\noindent \textbf{Supervised Pseudo-label Targets.} We include \textit{Depth Maps (Depth)} and \textit{Optical Flow (Flow)}. Intuitively, Depth and Flow can be considered as continuous pseudo ``labels'', which are made by models trained to perform depth and optical flow estimation~\cite{ranftl2021dpt, teed2020raft}. Depth does not improve over baseline with VTM+MLM. The nature of depth maps are to separate the foreground from the background, thus may guide the model to ignore information from the background, even when they are relevant for solving downstream VidL tasks (-0.1\% on TGIF-Frame, -1.6\% on DiDeMo Retrieval). Flow only focuses on the moving part between frames, while ignores the spatial details of static components, thus fail on more spatially-focused TGIF-Frame task (-0.5\%). We also find that the optical flow estimation model easily fails with sparse sampling strategy, which is widely adopted in VidL pre-training.\footnote{Please find visualization examples in the Appendix.}

\vspace{0.5ex}
\noindent \textbf{Supervised Visual Feature Targets.} We include continuous features extracted from the last layers of image classification model~\cite{liu2021swin} (\emph{i.e.}, \textit{Spatial-focused Image Features (SIF)}) and action recognition model~\cite{liu2022video-swin} (\emph{i.e.}, \textit{Temporal-aware Video Features (TVF)}). We consider regressing supervised features from Swin-B or VidSwin-L\footnote{VidSwin-L is trained on Kinetics-400~\cite{kay2017kinetics} with 83.1\% accuracy.} as a type of knowledge distillation from unimodal models to our model. SIF achieves significant improvement over baseline (+0.7\% on TGIF-Frame and +5.8\% on AveR for DiDeMo-Retrieval). In contrast, TVF fails to improve TGIF-Frame accuracy (-0.1\%), though it brings notable improvement on retrieval performance (+3.6\% on AveR). By distilling the knowledge from Swin-B, we enforce the model to focus more on spatial details of each frame, which we hypothesize is the main reason behind the large performance improvement. As previous study~\cite{buch2022revisiting} pointed out, existing VidL benchmarks largely test on spatial understanding about the key frame of the video, with only a fractional of examples actually testing on temporal reasoning over multiple frames. 

\vspace{0.5ex}
\noindent \textbf{Self-supervised Multimodal Feature Targets.} We use \textit{Discrete Visual Tokens (VQ)} from DALL-E~\cite{ramesh2021dalle} and continuous \textit{Multimodal Features (MMF)} extracted from CLIP~\cite{radford2021clip}. Both are pre-trained on large-scale image-text datasets, usually much more expensive than all other targets. Both targets improve the performance by a slight margin on only one task. VQ that can capture patch-level semantics, benefits TGIF-Frame (+0.3\%) which mostly focuses on scene understanding. While MMF from CLIP, contrastively pre-trained to measure the high-level similarity between the entire image and text sentence, is helpful for DiDeMo-Retrieval (+0.3\% on AveR). 

\vspace{0.5ex}
\noindent \textbf{Summary.} 
We hypothesize that many factors could lead to the low performance of an MVM target, such as its own characteristics (\textit{e.g.}, local vs. global features); the target model; the loss design; or the mismatch between pre-train objectives and downstream focus. We try our best to compare them rigorously with controlled experiments to find the best setting. Based on our experiments, regressing RGB values (Pixel) and distilling features from Swin-B~\cite{liu2021swin} (SIF) are the only two that produce consistent gains over the baseline on both downstream tasks. MVM with SIF achieves the best performance, with a gain of +0.7\% on TGIF-Frame and +5.8\% on AveR for DiDeMo-Retrieval over the baseline. Therefore, we use SIF as the default target for MVM in the following sections, unless specified otherwise.

\input{tabs/masking_strategy_webvid.tex}

\section{Analyses of MVM}
\noindent \textbf{Combining MVM Targets.} As different MVM targets focus on different aspects of visual modeling, a naive way to enable models with different visual capabilities is to combine them together. Specifically, the model pre-training can be supervised by more than one MVM loss, which is simply added together to be backpropagated. In Table~\ref{table:mvm-webvid-combine}, we find there is no merit in combining different MVM targets, leading to worse downstream performance than using SIF alone. When combining the best two targets found in Table~\ref{table:mvm-webvid}: Pixel+SIF, it performs better than Pixel only, but does not improve over using SIF alone. We hypothesize that the explicit details of pixel values may conflict with the high-level visual semantics summarized in the grid features from the image classifier. We further try to combine SIF with Flow in the hope of enforcing both temporal and spatial reasoning over video inputs. In addition, Flow is a better candidate than other targets, as it demonstrates some advantages on retrieval performance in Table~\ref{table:mvm-webvid}, and it is a different type of target from SIF, compared to temporal-aware video features.
The results are consistent, with improvements over optical flow only; while the performance drops, compared to SIF alone. Though our results are not encouraging, we believe how to effectively combine different MVM targets is an interesting direction for future study. 

\vspace{0.5ex}
\noindent \textbf{MVM Target Extractors \textit{vs.} Downstream Performance.} In Table~\ref{table:mvm-webvid-ic}, we explore different image classification models as the MVM target extractor for SIF, and investigate whether stronger image classification model enables better VidL performance. We compare ResNet-50~\cite{he2016resnet}, Swin-Tiny/Base/Large~\cite{liu2021swin},  trained on ImageNet-1K (IN1K) or ImageNet-22K (IN-22K)~\cite{deng2009imagenet}, and summarize the observations below:
\begin{itemize}[leftmargin=*, topsep=1pt]
    \setlength\itemsep{-2pt}
    \item ResNet-50 performs lower than Swin variants. Two potential reasons are ($i$) ResNet-50 architecture is very different from 
    VidSwin (\textit{i.e., with different inductive bias}); and ($ii$) the much lower ImageNet performance ($\sim$76 vs. $>$81) suggest the ResNet-50 features are not as strong. 
    \item When the target model shares \textit{similar inductive biases} to the video encoder (\textit{i.e.}, Swin-T/B/L), the downstream performance is \textit{not directly proportional to} ImageNet accuracy, and is overall better than that of Res50. This suggests that the architecture design of both target model and video encoder should be similar. 
    \item A key difference between different Swin targets is the feature dimension (768/1024/1568 for Swin-T/B/L), while the video tokens from CT are of size 768. Although we project them into the same dimension as the targets, the mismatch may lead to slightly lower performance (with Swin-L especially). 
\end{itemize}
In short, we believe a SIF target model should share similar inductive biases as the video encoder. 

\vspace{0.5ex}
\noindent \textbf{Masking Strategy.}
We investigate the effect of different masking strategies in Table~\ref{table:mvm-webvid-m_strategy}, including random masking (RM), blockwise masking (BM), attended masking (AM), and their combinations. 

\begin{itemize}[leftmargin=*, topsep=1pt]
    \setlength\itemsep{-3pt}
    \item \textbf{Random Masking (RM).} Following the conventional practice in MLM, we randomly select a certain percentage $p_m$ of video frame patches from the whole video inputs to be masked. In Table~\ref{table:mvm-webvid-m_ratio}, we explore different masking ratios ($p_m$), and empirically find $p_m = 30\%$ gives the best downstream performance.
    \item \textbf{Blockwise Masking (BM).} To make MVM relying less on similar neighbor patches, we adopt blockwise masking~\cite{tan2021vimpac,bao2022beit} that masks blocks of video patches along spatial-temporal dimension rather than independently masking randomly sampled patches for each frame. Specifically, we randomly sample an $(H', W', T')$ as a masking block, where all $H'\times W'$ visual patches in the following $T'$ consecutive frames will be masked; we repeat this process until $>p_m$ of video patches are masked to perform MVM pre-training. 
    \item \textbf{Attended Masking (AM).} Attended masking tries to put more weights on the more important elements based on the attention weights computed by Cross-modal Transformer (CT). A similar idea has been explored in~\cite{zellers2021merlot} for MLM. Here, we extend AM to both visual and textual modalities. We first keep the video-text inputs intact, feed them into CT to compute the attention weights, to decide which portions in video and text are more important. We then select the top $p_m$ of most-attended  patches/tokens to be masked in video-text inputs for MVM and MLM.
\end{itemize}

\input{tabs/mvm_target_on_cc.tex}

\input{tabs/mvm_on_webvid+cc.tex}

To combine different masking strategies, we randomly apply one masking method for each video-text pair in a batch. Results in Table~\ref{table:mvm-webvid-m_strategy} suggest that TGIF-Frame can slightly benefit from BM, and combining BM with RM leads to the best retrieval performance on DiDeMo. As video usually presents analogous visual patterns in spatial-temporal neighbors (\textit{i.e.}, nearby patches within current frame or neighboring frames), when masking patches independently (\textit{i.e.}, RM), these neighbors can make the masked patches easy to recover, and may lead to spurious success in MVM evaluation.  By masking a block (\textit{i.e.}, BM) instead of individual patches,  the model cannot merely rely on similar neighboring visual cues but requires actual visual reasoning to recover a group of missing patterns. Combining BM with RM leads to more diverse dropout patterns in video inputs, which is in analogy to data augmentation.

In addition, AM and combinations with AM are not effective for both downstream tasks. It is also worth noting that AM greatly increase the training time (4 times more than RM/BM), due to the additional forward pass needed to compute the attention weights. In our implementation, we optimize the three losses altogether in the same forward-backward pass. Hence, the performance drop with AM may be due to the important elements (\textit{e.g.}, visual patches containing the main object or content words) are more likely to be masked together and leaving the less relevant elements (\textit{e.g.}, scene background or stop words) intact, which will especially make the learning of video-text matching harder. 

\vspace{0.5ex}
\noindent \textbf{Applying MVM to Image-Text Data.}
As image can be considered as a special case of video with temporal size 1, video-language (VidL) pre-training can take advantages of image-text data, which has been proven successful in~\cite{lei2021clip-bert, bain2021frozen}. The current trend in VidL pre-training is to leverage both video-text data and image-text data. Therefore, we repeat the experiments in Section~\ref{sec:study} and examine which MVM targets work the best on downstream VidL tasks, when pre-trained on image-text data only. We remove optical flow and temporal-aware video features from this study, as the inputs are static images. In Table~\ref{table:mvm-cc}, we pre-train our model on CC3M~\cite{sharma2018cc} for 5 epochs and report results on TGIF-Frame and DiDeMo-Retrieval. The performance trend with different MVM targets are not consistent with that observed on video-text data. Pixel is able to largely preserve the baseline (VTM+MLM) performance, while other MVM targets lead to different degrees of performance drop, especially on retrieval. Without visual implications from neighbor frames as video, MVM is more challenging to learn on image data. On the other hand, MVM over an image may easily fit in static visual representation, which could hurt video temporal reasoning and not benefit downstream VidL learning.

\vspace{0.5ex}
\noindent \textbf{Combining Video-Text Data with Image-Text Data.}
We further follow recent VidL literature~\cite{bain2021frozen,li2022alpro} to use both video-text data and image-text data for pre-training, and investigate different ways to combine MVM targets on image and video data in Table~\ref{table:mvm-webvid+cc}. Note that we adopt the best training strategy found in the above investigations, that is, using spatial-focused image feature (SIF) as MVM target for video inputs, and using blockwise masking (BM) + random masking (RM) with masking ratio of 30\% as the masking strategy. As the best MVM target (Pixel) on image data does not show improvement over the baseline without MVM objective in Table~\ref{table:mvm-cc}, we explore with/without MVM objective on images in this combined pre-training. For the baseline with VTM+MLM only, we simply remove the MVM objective on both image and video data, while keeping the rest training settings. Under the strict fair comparison, we observe adding MVM objectives contributes to $>$+0.4\% gains on TGIF-Frame and $>$+2.6\% increase on AveR for DiDeMo-Retrieval.  
Comparing with or without MVM objective on images, adding MVM on image-text brings minor performance difference (+0.2\%  on TGIF-Frame and degrades by -0.3\%  on AveR for DiDeMo-Retrieval) over MVM on video-text only. 
Therefore, in our final setting, we only apply MVM objective on video data.

\input{tabs/qa_and_caption.tex}

\input{tabs/retrieval.tex}

\section{Main Results} \label{sec:sota}
To this end, we combine the most effective MVM strategies to pre-train \modelname and evaluate on 13 video-language (VidL) tasks. Table~\ref{table:qa} shows the comparison to prior arts on \textbf{video question answering (QA)} and \textbf{video captioning}. We observe that \modelname is effective in learning transferable knowledge for the downstream tasks. For example, considering pre-training data at a similar scale (\textit{i.e.,} $\leq$ 5M, the top rows of Table~\ref{table:qa}), \modelname achieves better results than prior arts, including ALPRO~\cite{li2022alpro}, ClipBERT~\cite{lei2021clip-bert}, and SwinBERT~\cite{lin2022swin-bert}, across all considered video QA and video captioning benchmarks. Specifically, when pre-training with the exact same data (\textit{i.e.}, WebVid2.5M~\cite{bain2021frozen} + CC3M~\cite{sharma2018cc}). \modelname surpasses ALPRO by 2.4\% accuracy on MSRVTT-QA and 8.4\% accuracy on MSVD-QA, respectively.
We also compare with other models pre-trained on significantly larger scale of video-text pairs. As shown in the bottom rows of Table~\ref{table:qa}, although we use less pre-training data than others, \modelname still achieves comparable or better performance. \footnotetext{The SIF target model Swin-B is trained on IN-22K.}

We observe similar findings on video captioning. On MSRVTT captioning, \modelname is only 2 points behind MV-GPT~\cite{seo2022mv-gpt} pre-trained with 53M video-text  pairs, which is 10 times larger than ours (5M). In addition, MV-GPT leverages ASR transcripts to enhance the captioning performance, while our captioning model takes only video frames as inputs and outputs the video caption.\footnote{Details about downstream finetuning on captioning are in Appendix.} 
We believe augmenting \modelname with additional modalities, such as audio or ASR transcripts, can further improve captioning performance, which we leave as future work.

Table~\ref{table:retrieval-all} presents the comparison on \textbf{text-to-video retrieval}. When pre-training with the same datasets (\textit{i.e.}, WebVid2.5M~\cite{bain2021frozen} + CC3M~\cite{sharma2018cc}), \modelname shows across-the-board improvements with all metrics considered on DiDeMo and LSMDC. It is worth noting that our method performs comparably to BridgeFormer~\cite{ge2022bridge-former} on MSRVTT-Retrieval. BridgeFormer adopts a noun/verb masking strategy during pre-training, which is specially aligned to the simple sentences in MSRVTT. However, it cannot show similar effects on DiDeMo and LSMDC due to more complex texts with multiple nouns/verbs. In contrast, the studied MVM can achieve a comprehensive enhancement in VidL learning and lead to notable improvements (+10.9\% R1 on DiDeMo and +6.1\% R1 on LSMDC).

\noindent\textbf{Direct Comparison to \modelorig~\cite{fu2021violet}}. Across Table~\ref{table:qa} and~\ref{table:retrieval-all}, it is worth noting that \modelname outperforms \modelorig with notable margins, even when \modelorig is pre-trained with significantly more data (about 37 times more). Specifically, \modelname yields an average gain of +3.4\% across 8 video QA datasets, and an absolute gain of +8.6\% on R1 across all three retrieval benchmarks. These results suggest the importance of an appropriate MVM setting, which is the core belief in our study.

%% file: tabs/combine_target.tex
% \begin{table}[t]
% \centering
%     \tablestyle{3pt}{1.1} 
%     \def \w{20pt} 
%     % \resizebox{\linewidth}{!}{
%     \begin{tabular}{l| ccccc}
%         \shline
%         \multirow{2}{*}{MVM Targets} & TGIF-Frame & \multicolumn{4}{c}{DiDeMo-Retrieval} \\
%         \cmidrule(lr){2-2}\cmidrule(lr){3-6}
%          & Acc. & R1 & R5 & R10 & AveR \\
%         \hline
%          Pixel & 68.3 & 29.2 & 58.6 & 70.1 & 52.6\\
%          Optical Flow & 67.6 & 30.3 & 58.0 & 70.3 & 52.9\\
%          \rowcolor{lightgray}
%           SIF & \textbf{68.8} & \textbf{35.1} & \textbf{63.3} & \textbf{73.1} & \textbf{57.2} \\
%         \hline
%         SIF + Pixel & \textbf{68.8} & 31.8 & 60.4 & 73.0 & 55.1\\

%         SIF + Optical Flow & 68.7 & 34.4 & 61.5 & 72.8 & 56.3\\
%         \shline
%     \end{tabular}
%     % }
%     \vspace{-2ex}
%     \caption{\textbf{Combining target features for MVM}. All
% variants are pre-trained on WebVid~\cite{bain2021frozen} for 5 epochs. The final pre-training setting is highlighted in \colorbox{lightgray}{gray}.}
%     \label{table:mvm-webvid-combine}
% \end{table}

\begin{table*}
  \begin{minipage}{0.85\columnwidth}
    \centering
    \tablestyle{5pt}{1.0} 
    % \resizebox{0.4\columnwidth}{!}{%
     \begin{tabular}{l| ccccc}
        \shline
        \multirow{2}{*}{MVM Targets} & TGIF-Frame & \multicolumn{4}{c}{DiDeMo-Retrieval} \\
        \cmidrule(lr){2-2}\cmidrule(lr){3-6}
         & Acc. & R1 & R5 & R10 & AveR \\
        \hline
         Pixel & 68.3 & 29.2 & 58.6 & 70.1 & 52.6\\
         Flow & 67.6 & 30.3 & 58.0 & 70.3 & 52.9\\
         \rowcolor{lightgray}
          SIF & \textbf{\siftgif} & \textbf{\sifdro} & \textbf{\sifdrf} & \textbf{\sifdrt} & \textbf{\sifaver} \\
        \hline
        SIF + Pixel & \textbf{68.8} & 31.8 & 60.4 & 73.0 & 55.1\\

        SIF + Flow & 68.7 & 34.4 & 61.5 & 72.8 & 56.3\\
        \shline
    \end{tabular}
    % }
    \caption{\textbf{Combining MVM targets}. All
variants are pre-trained on WebVid~\cite{bain2021frozen} for 5 epochs, using RM with 15\% as the masking strategy. We highlight the final setting in \colorbox{lightgray}{gray}. }
    \label{table:mvm-webvid-combine}
  \end{minipage}\hfill % maximize the horizontal separation
  \begin{minipage}{1.15\columnwidth}
    \centering
    \tablestyle{4pt}{1.0} 
    % \resizebox{\columnwidth}{!}{%
    \begin{tabular}{l cc|ccccc}
        \shline
        Image Features & Train & IN-1K & TGIF-Frame & \multicolumn{4}{c}{DiDeMo-Retrieval} \\
        \cmidrule(lr){3-3} \cmidrule(lr){4-4} \cmidrule(lr){5-8}
         Model & Data & ACC@1 & Acc. & R1 & R5 & R10 & AveR \\
        \hline
        ResNet-50~\cite{he2016resnet} & IN-1K & 76.1 & 67.3 & 29.1 & 58.1 & 69.3 & 52.2\\ 
        Swin-T~\cite{liu2021swin} & IN-1K & 81.2 & \textbf{68.9} & 33.8 & \textbf{63.6} & 74.2 & 57.2\\
        Swin-B & IN-1K & 83.5 & 68.3 & 34.9 & 63.4 & 73.9 & 57.4\\
        \rowcolor{lightgray}
         Swin-B & IN-22K & 85.2  & \siftgif & \textbf{\sifdro} & \sifdrf & \textbf{\sifdrt} & \textbf{\sifaver}\\
         % 68.8 & \textbf{35.1} & 63.3 & 73.1 & \textbf{57.2} \\
         Swin-L & IN-22K & \textbf{86.3}   & 68.2 & 33.2 & 62.4 & 72.6 & 56.1\\
        \shline
    \end{tabular}
    % }
    \caption{\textbf{Comparing different image feature targets for MVM.} All variants are pre-trained on WebVid~\cite{bain2021frozen} with VTM+MLM+MVM (SIF) for 5 epochs, using RM with 15\% as the masking strategy. The final pre-training setting is highlighted in \colorbox{lightgray}{gray}. 
    % \linjie{tuning default setting, hope to get better results to ease the discussion.}
    }
    \label{table:mvm-webvid-ic}
  \end{minipage}
  
\end{table*}

%% file: tabs/masking_strategy_webvid.tex
% \begin{table}[t]
% \centering
%     \tablestyle{3pt}{1.1} 
%     \def \w{20pt} 
%     \resizebox{\linewidth}{!}{
%     \begin{tabular}{lc | cccccc}
%         \shline
%         Masking  & Time Cost & TGIF-Frame & \multicolumn{4}{c}{DiDeMo-Retrieval}  \\
%         \cmidrule(lr){2-2}\cmidrule(lr){3-3}\cmidrule(lr){4-7}
%          Strategy & hours & Acc. & R1 & R5 & R10 & AveR \\
%         \hline
%          RM & 8.0 & 68.8 & 35.1 & 63.3 & 73.1 & 57.2\\
%          BM & 8.0 & \textbf{69.0} & 35.9 & 63.3 & 74.6 & 57.9\\
%           AM  &  34.5 & 68.4 & 31.5 & 59.9 & 72.0 & 54.7\\
%           % {'r@1': '31.46', 'r@5': '59.86', 'r@10': '72.01', 'median': '3'}
%           \hline
%           \rowcolor{lightgray}
%           RM+BM & 8.0 & 68.6 & \textbf{36.4} & \textbf{64.2} & 74.4 & \textbf{58.3}\\
%           RM+AM & 20.5 & 68.8 & 33.7 & 63.2 & 73.5 & 56.8\\
%           % {'r@1': '33.71', 'r@5': '63.23', 'r@10': '73.54', 'median': '3'}
%           BM+AM & 20.5 & 68.9 & 35.6 & 61.9 & 74.4 & 57.3 \\
%           \hline
%           RM+BM+AM  & 17.0 & 68.6 & 34.7 & 62.0 & \textbf{74.8} & 57.2\\
%         \shline
%     \end{tabular}
%     }
%     \vspace{-2ex}
%     \caption{Impact of \textbf{masking strategy for MVM}. All variants are pre-trained on WebVid~\cite{bain2021frozen} with VTM+MLM+MVM (2d feature) for 5 epochs. The masking ratio is set as 15\% for all masking strategies. The final pre-training setting is highlighted in \colorbox{lightgray}{gray}.}
%     \label{table:mvm-webvid-m_strategy}
% \end{table}

\begin{table*}
  \begin{minipage}{1.2\columnwidth}
    \centering
    \tablestyle{7pt}{1.0} 
    % \resizebox{0.4\columnwidth}{!}{%
     \begin{tabular}{lc | cccccc}
        \shline
        Masking  & Time Cost & TGIF-Frame & \multicolumn{4}{c}{DiDeMo-Retrieval}  \\
        \cmidrule(lr){2-2}\cmidrule(lr){3-3}\cmidrule(lr){4-7}
         Strategy & hours & Acc. & R1 & R5 & R10 & AveR \\
        \hline
         RM & 8.0 & \siftgif & \sifdro & \sifdrf & \sifdrt & \sifaver\\
         BM & 8.0 & \textbf{69.0} & 35.9 & 63.3 & 74.6 & 57.9\\
          AM  &  34.5 & 68.4 & 31.5 & 59.9 & 72.0 & 54.7\\
          % {'r@1': '31.46', 'r@5': '59.86', 'r@10': '72.01', 'median': '3'}
          \hline
          \rowcolor{lightgray}
          RM+BM & 8.0 & 68.7 & \textbf{36.4} & \textbf{64.2} & 74.4 & \textbf{58.3}\\
          RM+AM & 20.5 & 68.8 & 33.7 & 63.2 & 73.5 & 56.8\\
          % {'r@1': '33.71', 'r@5': '63.23', 'r@10': '73.54', 'median': '3'}
          BM+AM & 20.5 & 68.9 & 35.6 & 61.9 & 74.4 & 57.3 \\
          \hline
          RM+BM+AM  & 17.0 & 68.6 & 34.7 & 62.0 & \textbf{74.8} & 57.2\\
        \shline
    \end{tabular}
    % }
    \caption{Impact of \textbf{masking strategy of MVM}. All variants are pre-trained on WebVid~\cite{bain2021frozen} with VTM+MLM+MVM (SIF) for 5 epochs. The masking ratio is set as 15\% for all masking strategies. The final pre-training setting is highlighted in \colorbox{lightgray}{gray}.}
    \label{table:mvm-webvid-m_strategy}
  \end{minipage}\hfill % maximize the horizontal separation
  \begin{minipage}{0.8\columnwidth}
    \centering
    \tablestyle{6pt}{1.1} 
    % \resizebox{\columnwidth}{!}{%
    \begin{tabular}{c| ccccc}
        \shline
        $p_m$ & TGIF-Frame & \multicolumn{4}{c}{DiDeMo-Retrieval} \\
        \cmidrule(lr){2-2}\cmidrule(lr){3-6}
         & Acc. & R1 & R5 & R10 & AveR \\
        \hline
         15\% & \siftgif & \sifdro & \sifdrf & \sifdrt & \sifaver\\
         % 0.20 & 68.7 & \\
         % {'r@1': '35.55', 'r@5': '63.74', 'r@10': '74.77', 'median': '3'}
         \rowcolor{lightgray}
          30\%  &  68.8 & \textbf{36.2} & \textbf{64.0} & \textbf{74.5} & \textbf{58.2}\\
          % 0.40 & \\
          45\% & \textbf{68.9} & 35.6 & 61.9 & 74.4 & 57.3 \\
          % 0.50  & 68.9 & 36.6 & 64.0 & 74.1 & 58.2\\
          % {'r@1': '36.57', 'r@5': '64.04', 'r@10': '74.06', 'median': '3'}\\
          60\%  & 68.1 & 34.1 & 63.9 & 74.6 & 57.5\\
          % {'r@1': '34.12', 'r@5': '63.94', 'r@10': '74.57', 'median': '3'}
          % 0.70 & 68.3 & 
          % {'r@1': '35.75', 'r@5': '62.72', 'r@10': '73.85', 'median': '3'}
          75\%  & 68.3 & 35.4 & 62.4 & 74.2 & 57.3 \\
          % {'r@1': '35.44', 'r@5': '62.41', 'r@10': '74.16', 'median': '3'}
        \shline
         % \multicolumn{2}{c}{}&\\
         % \multicolumn{2}{c}{}&\\
    \end{tabular}
    % }
    \vspace{3pt}
    % \vspace{-3pt}
    \caption{Impact of \textbf{masking ratio of MVM}. All
variants are pre-trained on WebVid~\cite{bain2021frozen} with VTM+MLM+MVM (SIF) for 5 epochs, using RM as the masking strategy. The final pre-training setting is highlighted in \colorbox{lightgray}{gray}.}

    \label{table:mvm-webvid-m_ratio}
  \end{minipage}
\end{table*}

%% file: tabs/mvm_target_on_cc.tex
% \begin{table}[t]
% \centering
%     \tablestyle{3pt}{1.1} 
%     \def \w{20pt} 
%     % \resizebox{\linewidth}{!}{
%     \begin{tabular}{ll| ccccc}
%         \shline
%         Pre-training & \multirow{2}{*}{MVM Target} & TGIF-Frame & \multicolumn{4}{c}{DiDeMo-Retrieval} \\
%         \cmidrule(lr){3-3}\cmidrule(lr){4-7}
%         Task &  & Acc. & R1 & R5 & R10 & AveR \\
%         \hline
%         % None & None &  \\
%         ITM+MLM & None &  \textbf{69.8} &  \textbf{36.4} & 64.3 & 74.7 & \textbf{58.4}\\
%         \hline
%         \multirow{8}{*}{+MVM} & Pixel & 69.7 & 35.8 & \textbf{64.4} & \textbf{74.9} & \textbf{58.4} \\
%           & HOG~\cite{dalal2005hog} & \textbf{69.8} & 34.9 & \textbf{64.4} & 75.1 & 58.1 \\
%          \cline{2-7}
%          & VQ Token (DALL-E~\cite{ramesh2021dalle}) & \textbf{69.8} &  34.4 & 62.6 & 75.1 & 57.4  \\
%           & Multimodal tokens (CLIP~\cite{radford2021clip}) & \textbf{69.8} & 33.6 & 62.9 & 75.6 & 57.4\\
%          \cline{2-7} 
%         & Depth (DPT~\cite{Ranftl2021dpt}) & 69.6 & 32.3 & 63.8 & 74.2 & 56.9 \\
%           & 2D Feature (Swin~\cite{liu2021swin}) & 69.7 & 31.6 & 60.5 & 72.5 & 54.9 \\
%           & 3D Feature (VidSwin~\cite{liu2022ideo-swin}) & 68.9 & 30.8 & 60.4 & 72.1 & 54.4  \\
%         \shline
%     \end{tabular}
%     % }
%     \caption{\textbf{Comparing target features for MVM (video)}. All
% variants are pre-trained on CC3M~\cite{sharma2018cc} for 5 epochs.}
%     \vspace{-2ex}
%     \label{table:mvm-cc}
% \end{table}

\begin{table*}[t]
\centering
    \tablestyle{7pt}{1.0} 
    \def \w{20pt} 
    % \resizebox{\linewidth}{!}{
    \begin{tabular}{ll| ccccc}
        \shline
        \multirow{2}{*}{Pre-training Tasks}  & \multirow{2}{*}{MVM Target} & TGIF-Frame & \multicolumn{4}{c}{DiDeMo-Retrieval} \\
        \cmidrule(lr){3-3}\cmidrule(lr){4-7}
        &  & Acc. & R1 & R5 & R10 & AveR \\
        \hline
        % None & None &  \\
        ITM+MLM & None &  \textbf{69.8} &  \textbf{36.4} & 64.3 & 74.7 & \textbf{58.4}\\
        \hline
        \multirow{6}{*}{+MVM} & RGB Pixel Values & 69.7 {\color{BrickRed}(-0.1)} & 35.8 {\color{BrickRed}(-0.6)} & \textbf{64.4 {\color{ForestGreen}(+0.1)}} & 74.9 {\color{ForestGreen}(+0.2)} & \textbf{58.4} \\
          & Histogram of Oriented Gradients~\cite{dalal2005hog} & \textbf{69.8} & 34.9 {\color{BrickRed}(-1.5)} & \textbf{64.4 {\color{ForestGreen}(+0.1)}} & 75.1 {\color{ForestGreen}(+0.4)} & 58.1 {\color{BrickRed}(-0.3)}\\
         \cline{2-7}
        & Depth Maps (DPT-L~\cite{ranftl2021dpt}) & 69.6 {\color{BrickRed}(-0.2)} & 32.3 {\color{BrickRed}(-4.1)} & 63.8 {\color{BrickRed}(-0.5)} & 74.2 {\color{BrickRed}(-0.5)} & 56.9 {\color{BrickRed}(-1.5)}\\
         \cline{2-7} 
          & Spatial-focused Image Features (Swin-B~\cite{liu2021swin}) & 69.7 {\color{BrickRed}(-0.1)} & 31.6 {\color{BrickRed}(-4.8)} & 60.5 {\color{BrickRed}(-3.8)} & 72.5 {\color{BrickRed}(-2.2)} & 54.9 {\color{BrickRed}(-3.5)}\\
          % & Temporal-aware Video Feature (VidSwin-B~\cite{liu2022ideo-swin}) & 68.9 {\color{BrickRed}(-0.9)} & 30.8 {\color{BrickRed}(-5.6)} & 60.4 {\color{BrickRed}(-3.9)} & 72.1 {\color{BrickRed}(-2.6)} & 54.4 {\color{BrickRed}(-4.0)}\\
         \cline{2-7} 
         & Discrete Visual Tokens (DALL-E~\cite{ramesh2021dalle}) & \textbf{69.8} &  34.4 {\color{BrickRed}(-2.0)} & 62.6 {\color{BrickRed}(-1.7)} & 75.1 {\color{ForestGreen}(+0.4)} & 57.4 {\color{BrickRed}(-1.0)}\\
          & Multimodal Features (CLIP-ViT-B~\cite{radford2021clip}) & \textbf{69.8} & 33.6 {\color{BrickRed}(-2.8)} & 62.9 {\color{BrickRed}(-1.4)} & \textbf{75.6 {\color{ForestGreen}(+0.9)}} & 57.4 {\color{BrickRed}(-1.0)}\\
        \shline
    \end{tabular}
    % }
    \caption{\textbf{Comparing target features for MVM applied to image-text data}. All
variants are pre-trained on CC3M~\cite{sharma2018cc} for 5 epochs. Masking is performed randomly (RM) with ratio of 15\%.}
\vspace{3pt}
    \label{table:mvm-cc}
\end{table*}

%% file: tabs/mvm_on_webvid+cc.tex
\begin{table*}[t]
\centering
    \tablestyle{10pt}{1.1} 
    \def \w{20pt} 
    % \resizebox{\linewidth}{!}{
    \begin{tabular}{l|cc|ccccc}
        \shline
        \multirow{2}{*}{Pre-training Tasks} & \multicolumn{2}{c|}{MVM Target} & TGIF-Frame & \multicolumn{4}{c}{DiDeMo-Retrieval} \\
        \cmidrule(lr){2-3}\cmidrule(lr){4-4}\cmidrule(lr){5-8}
         &  WebVid2.5M & CC3M & Acc. & R1 & R5 & R10 & AveR \\
        \hline
        % None & None &  \\
        VTM+MLM & None & None & 69.7  &  36.7 & 66.5 & 76.6 & 59.9\\
        \hline
        \rowcolor{lightgray}
        +MVM & Spatial-focused Image Features (Swin-B~\cite{liu2021swin}) & None & 71.1 & 38.8 & \textbf{69.6} & \textbf{80.0} & \textbf{62.8}\\
        % ep 5  {'r@1': '44.84', 'r@5': '73.95', 'r@10': '82.43', 'median': '2'}
        & Spatial-focused Image Features (Swin-B) & Pixel & \textbf{71.3} & \textbf{39.7} & 69.3 & 78.4 & 62.5\\
        % ep 5 {'r@1': '45.05', 'r@5': '73.75', 'r@10': '82.53', 'median': '2'}
        \shline
    \end{tabular}
    % }
    \caption{\textbf{Combining MVM target features for both video-text and image-text data}. All
variants are pre-trained on WebVid2.5M~\cite{bain2021frozen} +CC3M~\cite{sharma2018cc} for 5 epochs. 
% We adopt the best masking strategy of BM+RM, with masking ratio of 30\%. 
The final pre-training setting is highlighted in \colorbox{lightgray}{gray}.}
    \label{table:mvm-webvid+cc}
\end{table*}

%% file: tabs/qa_and_caption.tex
\begin{table*}[t]
\centering
    \tablestyle{7pt}{1.1} 
    % \resizebox{\linewidth}{!}{
    \begin{tabular}{ll|cccccccc|cc}
        % \toprule
        \shline
        ~ & \# Pretrain  & \multicolumn{3}{c}{TGIF~\cite{jang2017tgif-qa}}  & \multicolumn{2}{c}{MSRVTT\cite{xu2016msrvtt}}  & \multicolumn{2}{c}{LSMDC~\cite{torabi2016lsmdc-fib}}  & MSVD~\cite{chen2011msvd}  & \multicolumn{2}{c}{Captioning}\\
        \cmidrule(lr){3-5} \cmidrule(lr){6-7} \cmidrule(lr){8-9} \cmidrule(lr){10-10} \cmidrule(lr){11-12}
        Method & videos/images & Act. & Trans. & Frame  & MC~\cite{yu2018js-fusion} & QA~\cite{xu2017msrvtt-qa}  & MC & FiB  & QA~\cite{xu2017msrvtt-qa} & MSRVTT  & MSVD \\
        \shline
        ClipBERT~\cite{lei2021clip-bert} & 0.2M & 82.8 & 87.8 & 60.3  & 88.2 & 37.4  & - & -  & -  & -  & -\\
        ALPRO~\cite{li2022alpro} & 5M & - & - & - & - & 42.1 & - & - & 46.3 & - & - \\
        SwinBERT~\cite{lin2022swin-bert} & - & - & - & -  & - & -  & - & -  & -  & 53.8  & 120.6\\
        
         \hline
         \multicolumn{5}{l}{\demph{\textit{Models pre-trained on more data}}}\\
         \hline
        \demph{JustAsk~\cite{yang2021just-ask}}& \demph{69M}  & \demph{-} & \demph{-} & \demph{-}  & \demph{-} & \demph{41.5}  & \demph{-} & \demph{-}  & \demph{46.3}  & \demph{- } & \demph{-}\\
        \demph{MERLOT~\cite{zellers2021merlot}} & \demph{180M}  &   \demph{94.0} & \demph{96.2} & \demph{69.5}  & \demph{90.9} & \demph{43.1}  & \demph{81.7} & \demph{52.9}  & \demph{-}  & \demph{-}  & \demph{-}\\
        \demph{All-in-one~\cite{wang2022all-in-one}} & \demph{283M} & \demph{95.5} & \demph{94.7} & \demph{66.3}  & \demph{92.3} &  \demph{46.8} & \demph{84.4} & \demph{-}  & \demph{48.3}  & \demph{-}  & \demph{-}\\
        \demph{MV-GPT~\cite{seo2022mv-gpt}} & \demph{53M}  & \demph{-} & \demph{-} & \demph{-}  & \demph{-} & \demph{41.7} & \demph{-} & \demph{-}  & \demph{-}  & \demph{60.0}  & \demph{-}\\
        \demph{\modelorig~\cite{fu2021violet}} & \demph{186M}  & \demph{92.5} & \demph{95.7} & \demph{68.9}  & \demph{91.9} & \demph{43.9} & \demph{82.8} & \demph{53.7}  & \demph{47.9}  & \demph{-}  & \demph{-}\\
         \hline
          \modelname & 5M\footnotemark  & \textbf{94.8} & \textbf{99.0} & \textbf{72.8}  & \textbf{97.6} & \textbf{44.5}  & \textbf{84.4} & \textbf{56.9}  & \textbf{54.7}  &  \textbf{58.0}  &  \textbf{139.2}\\
          %  & & 85.7 & 91.0
         \shline
        % \bottomrule
    \end{tabular}
    % }
    \caption{Comparison with SOTA on \textbf{video question answering} (accuracy) and \textbf{video captioning} (CIDEr). 
  \modelname is pre-trained on WebVid2.5M~\cite{bain2021frozen}+CC3M~\cite{sharma2018cc} with VTM+MLM+MVM (SIF on videos) for 10 epochs. 
  We gray out methods that use significantly more pre-training data. 
    % The 69M corpus is the 69M video questions constructed with HowTo100M videos in~\cite{yang2021just-ask}, 180M refers to the 180M YouTube clip-text pairs in YT-Temporal-180M~\cite{zellers2021merlot},  283 M is the combination of WebVid and YT-Temporal-180M, and the 53M corpus is also reformat of HowTo100M, designed for multi-modal captioning pre-training in~\cite{seo2022mv-gpt}.
% Note that MV-GPT~\cite{seo2022mv-gpt} additionally leverages ASR transcripts as inputs. 
    }
    \vspace{3pt}
    \label{table:qa}
\end{table*}

%% file: tabs/retrieval.tex
\begin{table*}[t!]
\centering
    \tablestyle{11pt}{1.1} 
    % \resizebox{\linewidth}{!}{
    \begin{tabular}{ll|ccccccccc}
        % \toprule
        \shline
        ~ & \# Pretrain & \multicolumn{3}{c}{MSRVTT~\cite{xu2016msrvtt}}& \multicolumn{3}{c}{DiDeMo~\cite{hendricks2017didemo}} & \multicolumn{3}{c}{LSMDC~\cite{rohrbach2015lsmdc}}\\
        \cmidrule(lr){3-5} \cmidrule(lr){6-8} \cmidrule(lr){9-11}
        Method & videos/images & R1 & R5 & R10 & R1 & R5 & R10 & R1 & R5 & R10 \\
        \shline
        ClipBERT~\cite{lei2021clip-bert} & 0.2M & 22.0 & 46.8 & 59.9 & 20.4 & 48.0 & 60.8  & - & - & -\\
        Frozen~\cite{bain2021frozen} & 5M  & 31.0 & 59.5 & 70.5 & 31.0 & 59.8 & 72.4 & 15.0 & 30.8 & 39.8 \\
        ALPRO~\cite{li2022alpro} & 5M & 33.9 & 60.7 & 73.2 & 35.9 & 67.5 & 78.8   & - & - & -\\
        BridgeFormer~\cite{ge2022bridge-former} & 5M  & \textbf{37.6} & \textbf{64.8} & 75.1 & 37.0 & 62.2 & 73.9  & 17.9 & 35.4 & 44.5 \\
         \hline
         \multicolumn{5}{l}{\demph{\textit{Models pre-trained on more data}}}\\
         \hline
        \demph{HERO~\cite{li2020hero}} & \demph{136M} & \demph{16.8} & \demph{43.4} & \demph{57.7} & \demph{-} & \demph{-} & \demph{-} & \demph{-}  & \demph{-} & \demph{-}\\
        % \rowcolor{lightgray}
        % \demph{VideoCLIP~\cite{xu2021video-clip}} & \demph{136M} & \demph{30.9} & \demph{55.4} & \demph{66.8} & \demph{-} & \demph{-} & \demph{-} & \demph{-} & \demph{-} & \demph{-}\\
        % \rowcolor{lightgray}
        \demph{All-in-one~\cite{wang2022all-in-one}} & \demph{138M}  & \demph{37.9} & \demph{68.1} & \demph{77.1} & \demph{32.7} & \demph{61.4} & \demph{73.5}  & \demph{-} & \demph{-} & \demph{-} \\
        % \rowcolor{lightgray}
        % \demph{BridgeFormer~\cite{ge2022bridge-former}} & \demph{400M} & \demph{44.9 / 71.9 / 80.3} & \demph{-}  & \demph{21.8 / 41.1 / 50.6} \\
        % BLIP~\cite{li2022blip}  & - / 129M & 43.3 / 65.6 / 74.7 & - & - & - \\
        % \rowcolor{lightgray}
        \demph{Clip4Clip~\cite{luo2021clip4clip}} & \demph{400M} &  \demph{42.1} & \demph{71.9} & \demph{81.4} & \demph{43.4} & \demph{70.2} & \demph{80.6}  & \demph{21.6} & \demph{41.8} & \demph{49.8}\\
        \demph{\modelorig~\cite{fu2021violet}} & \demph{186M} &  \demph{34.5} & \demph{63.0} & \demph{73.4} & \demph{32.6} & \demph{62.8} & \demph{74.7}  & \demph{16.1} & \demph{36.6} & \demph{41.2}\\
        % \rowcolor{LightCyan}
        % CAMoE~\cite{}&\textcolor{LightCyan}{00.}-\textcolor{LightCyan}{M} /  \textcolor{LightCyan}{.}400M & 47.3 / 74.2 / 84.5 & 43.8 / 71.4 / \textcolor{LightCyan}{0.}-\textcolor{LightCyan}{0} & 49.8 / 79.2 / 87.0 & 25.9 / 46.1 / 53.7\\
        \hline
        \modelname & 5M & 37.2 & \textbf{64.8} & \textbf{75.8} & \textbf{47.9} & \textbf{76.5} & \textbf{84.1} & \textbf{24.0} & \textbf{43.5} & \textbf{54.1} \\
        \shline
        % \bottomrule
    \end{tabular}
    % }
    \caption{Comparison with SOTA on 
    \textbf{text-to-video retrieval} tasks (R1/5/10). 
    % All results are reported on R1/5/10.
    We gray out methods that use significantly more pre-training data. 
    % The 136M corpus is from HowTo100M~\cite{miech2019howto100m}, 0.2M refers to COCO~\cite{chen2015coco}+VG~\cite{krishna2017vg} data, 138M is the combination of HowTo100M and WebVid, 400M is the private image-text data used in CLIP~\cite{radford2021clip}.
    }
    \label{table:retrieval-all}
\end{table*}

%% file: appendix.tex
\appendix

\section{Additional Results}
\noindent \textbf{Fair Comparison to \modelorig~\cite{fu2021violet}.} 
% Our model is based on the video-language transformer proposed in~\cite{fu2021violet}. 
\modelorig proposes to augment VTM+MLM with masked visual token modeling for VidL pre-training, while only showing marginal improvements on downstream performance. In contrast, we conduct comprehensive investigations across different MVM targets and masking strategies to demonstrate that effective MVM training can largely improve downstream performance. Note that our study already encompasses the design of MVM in~\cite{fu2021violet}, that is MVM with VQ target and BM+AM as the masking strategy. To make a fair comparison between~\cite{fu2021violet} and our best setting (MVM with SIF target and BM+AM as the masking strategy), we reproduce~\cite{fu2021violet} under the same setting and report downstream performance in Table~\ref{table:violet-v1-v2}. Results show that our setting obtains a significant improvement, with +2.3\% on TGIF-Frame and +13.3\% on AveR for DiDeMo-Retrieval, respectively. These results suggest the importance of an appropriate MVM setting, which is the core belief in our study.
\input{tabs/violet_v1_vs_v2.tex}

\vspace{0.5ex}
\noindent \textbf{MVM \textit{vs.} Temporal self-supervised objectives.} In addition to the reconstructive MVM task, other self-supervised video modeling tasks can be explored, for example, Frame Order Modeling (FOM)~\cite{li2020hero,zellers2021merlot}, which reconstructs the temporal orders of shuffled frame inputs. In Table~\ref{table:fom}, we compare MVM (SIF) with FOM in ~\cite{zellers2021merlot}, when pre-trained on WebVid~\cite{bain2021frozen}. MVM (SIF) still leads to better performance, with a gain of +0.7\% on TGIF-Frame and +4.8\% on AveR for DiDeMo-Retrieval, respectively.
\input{tabs/supp_fom.tex}

\vspace{0.5ex}
\noindent \textbf{Initialization and Learning of Video Backbone.}
We investigate the effect of different initialized video backbones with or without MVM on VL inputs in Table~\ref{table:mvm-webvid-init}. At first, although the used video transformer (VT) is randomly initialized, the MVM training still enhances the visual representation and benefits the downstream video-language (VidL) tasks. Furthermore, MVM can also boost better initialized VT from VidSwin-B and lead to a comprehensive increase. Specifically, the improvement gap is more significant than random initialization, where we can learn better from MVM and enlarge its effectiveness during pre-training. We additionally compare two self-supervised initializations with MVM on video-only inputs, one with TVF as the MVM target and the other with SIF. Though VidL pre-training with MVM from supervised VidSwin-B initialization leads to the best downstream performance, we observe consistent performance improvement from MVM on VL inputs regardless of the initialization setting.
\input{tabs/supp_vid_init.tex}

\vspace{0.5ex}
\noindent \textbf{Type of MVM Loss.} We compare the type of loss function for the MVM training by using least absolute deviations ($l_1$) or least square errors ($l_2$) in Table~\ref{table:mvm-webvid-loss}. It is well known that the $l_1$ loss can be resistant to outlier data. We show that MVM through $l_1$ is also more robust and leads to better performance on both video question answering and text-to-video retrieval than the $l_2$ loss.
\input{tabs/supp_mvm_loss.tex}

\vspace{0.5ex}
\noindent \textbf{MVM Prediction Head.} We investigate the prediction head for MVM in Table~\ref{table:mvm-webvid-head}. As a result, a single linear layer is not enough to model the complicated distilling MVM features. (\textit{e.g.,} 31.3 \textit{vs.} 35.4 R1 on DiDeMo-Retrieval)
Therefore, we follow VTM and MLM to use 2-layer MLP as the prediction head for MVM.
\input{tabs/supp_mvm_head.tex}

\vspace{0.5ex}
\noindent \textbf{TVF Target Extractors vs. Downstream Performance.} We compare distilling video features from  VidSwin-B vs. VidSwin-L (the default setting in the main text) in Table~\ref{table:tvf}. Here, for experiments with VidSwin-B, the same VidSwin-B weight is used to initialize the video backbone and to extract the MVM target. Hence, the MVM objective can be easily minimized by simply ignoring the text inputs, which conflict with the other objectives. This variant is in principle similar to masked frame modeling in HERO~\cite{li2020hero}, the key difference lies in whether the video backbone is refined during pre-training. In addition, we investigate whether the sparse sampling of video frames when extracting TVF target is the key reason behind the lower performance of TVF, compared to SIF. Hence, we compare the default sparse sampling of 5 frames, against a dense-version of TVF target (feeding 16 frames into VidSwin-L). While the dense input is slightly beneficial, SIF still performs better, with absolute advantages of +0.4\% on TGIF-Frame and +1.8\% on AveR for DiDeMo-Retrieval.
\input{tabs/supp_tvf.tex}

\vspace{0.5ex}
\noindent \textbf{Additional Exploration in Combining MVM Targets.}
We explore the additional combination of distilling MVM targets in Table~\ref{table:mvm-webvid-combine-sif-tvf}. MVM with SIF has an obvious advantage over TVF only on both video question answering and text-to-video retrieval. While, considering SIF+TVF seems not to bring a robust improvement, especially decreasing text-to-video retrieval. The previous study~\cite{buch2022revisiting} shows that current VidL benchmarks primarily focus on spatial understanding of the key frame from videos. Furthermore, combining TVF with SIF results in excessive training overhead. Accordingly, we choose SIF as our final pre-training setting.
\input{tabs/supp_combine_target.tex}

\vspace{0.5ex}
\noindent \textbf{Additional Exploration of SIF Target.}
We explore a more advanced SIF target, DeiT~\cite{touvron2021training}, in Table~\ref{table:mvm-sif-deit}. These results show that Swin-B still has an advantage (a noticeable higher 34.9 R1 on retrieval), consistent with our previous observations in the main text. That is, SIF should share similar inductive biases to the video encoder (i.e., Swin-T/B/L).
\input{tabs/supp_deit.tex}

\vspace{0.5ex}
\noindent \textbf{Investigation of Training Recipe with CLIP Target.}
We presented initial results for varying training settings using CLIP/Swin-B targets and compare them with the default setting in Table 1. Swin-B had a significant advantage over CLIP. As we adjust the training recipe with CLIP target in Table~\ref{table:mvm-clip}, a better training recipe reduces the performance gap between the SIF target and the CLIP target. The results in turn suggest the importance of an effective MVM strategy (e.g., masking ratio). Though impossible to iterate over all settings, Swin-B remains competitive under the same training recipe, especially with limited training data (IN-22K vs. 400M). Note that we use CLIP image features as the MVM target, while other related works~\cite{ni2022expanding,lin2022frozen} use them as model inputs. One potential enhancement is to leverage the multimodal information from both image and text encoders in CLIP (similar to the use of both in~\cite{ni2022expanding}), which is an interesting direction to explore in future studies. However, our setup is still valid, as we aim to train a fusion-encoder architecture rather than a dual-encoder architecture as CLIP.
\input{tabs/supp_clip}

\input{tabs/supp_extended_table1.tex}
\begin{figure*}[t!]
    \centering
    \begin{subfigure}[t]{0.48\textwidth}
        \centering
        \includegraphics[width=0.8\linewidth]{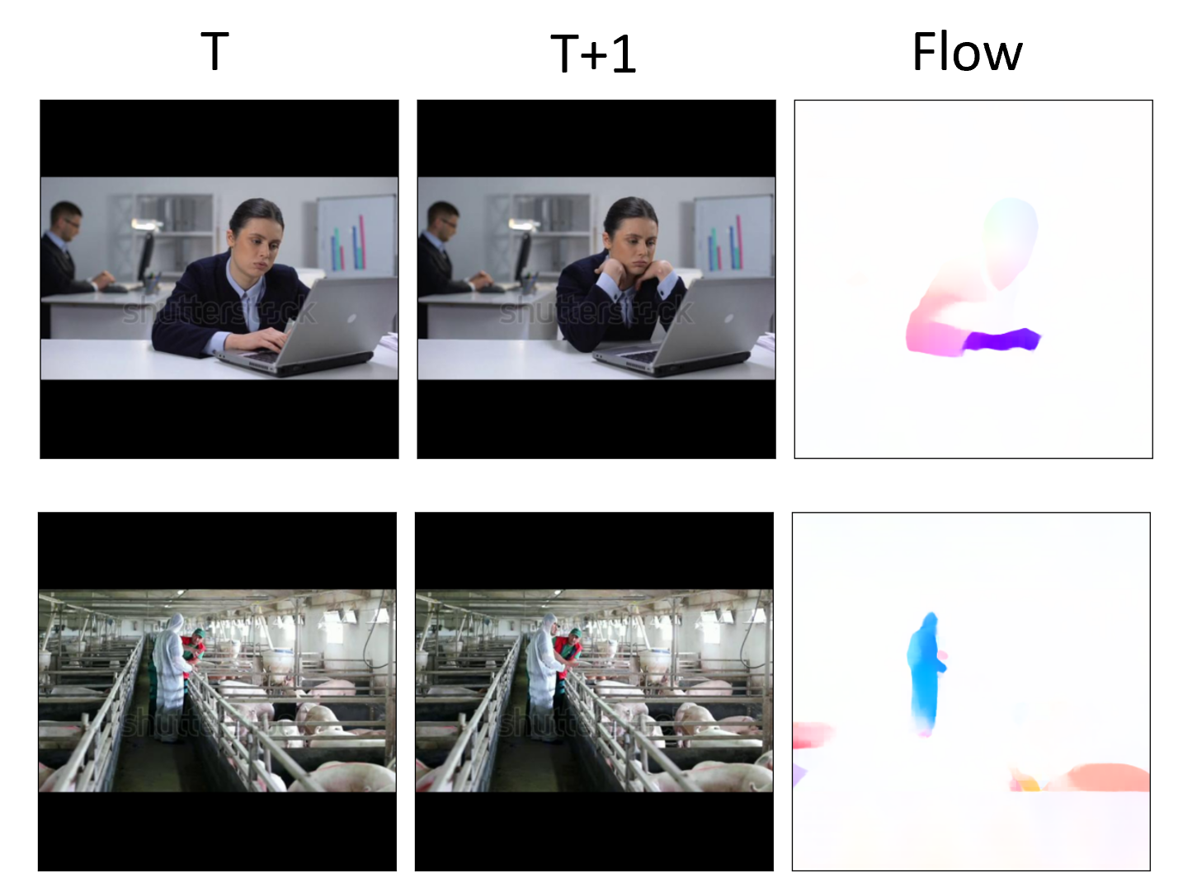}
        \caption{}
        \label{subfig:flow_good}
    \end{subfigure}
    \begin{subfigure}[t]{0.48\textwidth}
        \centering
        \includegraphics[width=0.8\linewidth]{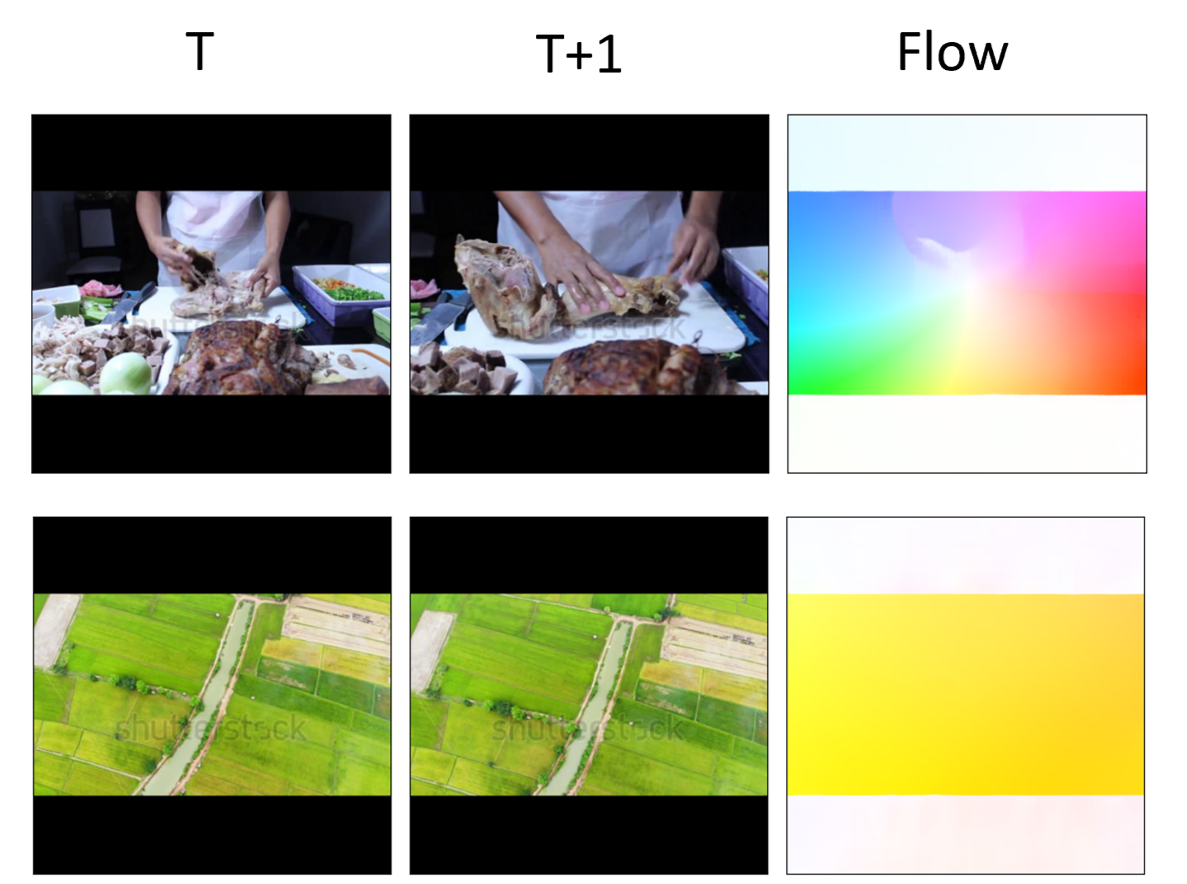}
        \caption{}
        \label{subfig:flow_bad}
    \end{subfigure}
    \vspace{13pt}
    \caption{\textbf{Visualization of optical flow (Flow) predictions by RAFT-L~\cite{teed2020raft} with sparsely sampled frames}. We show examples of good cases in (a) and bad cases in (b).}
    \label{fig:supp_vis_flow}
\end{figure*}

\vspace{0.5ex}
\noindent \textbf{Extended Results for Table 1.} The additional results on MSVD-QA and MSRVTT-Retrieval in Table~\ref{table:mvm-webvid-supp} show that \textit{SIF} is still the most effective, consistent with Table 1. However, the effects of different MVM targets seem to be better on MSVD-QA and MSRVTT-Retrieval (on average 15s long) than those on TGIF-Frame (on average 3s long) and DiDeMo-Retrieval (on average 30s long). We hypothesize that different downstream video lengths may contribute to different performance gains/losses when evaluating the effectiveness of an MVM target, which we leave as future directions for investigation. 

\vspace{0.5ex}
\noindent \textbf{Qualitative Results.}
Figure~\ref{fig:supp_vis_flow} shows good and bad examples of optical flow predictions made by RAFT-L~\cite{teed2020raft} with sparsely sampled frames. As shown in~\ref{subfig:flow_bad}, the top example shows zoom-in shots, and the bottom shows moving shots. All content in the current frame is moving, which is the main reason behind the failure in optical flow estimation.

\begin{figure*}[t]
\centering
    \includegraphics[width=\linewidth]{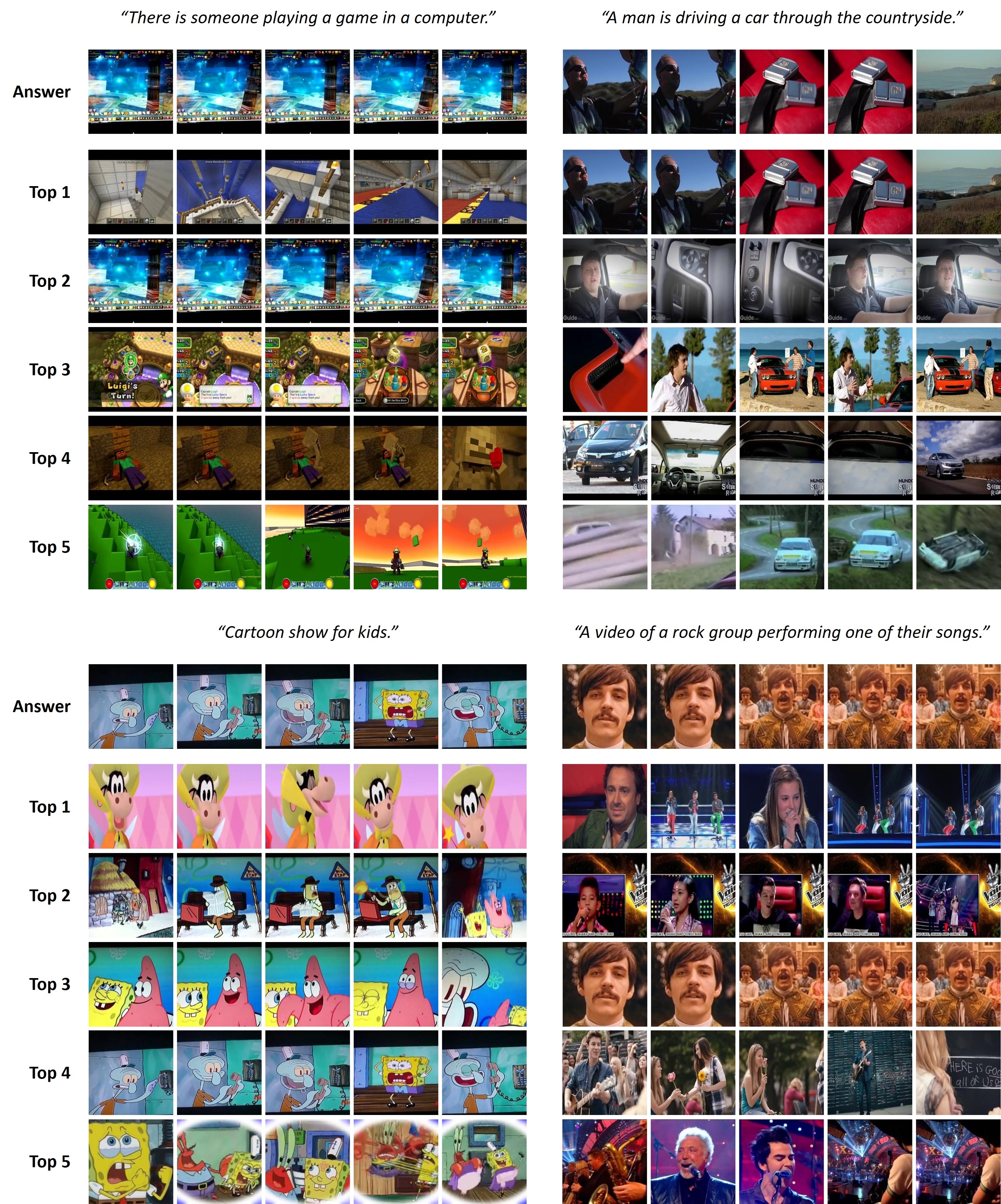}
    \caption{Qualitative examples of \textbf{zero-shot text-to-video retrieval} on MSRVTT~\cite{xu2016msrvtt}.}
    \label{fig:zs-msrvtt}
\end{figure*}

\begin{figure*}[t]
\centering
    \includegraphics[width=\linewidth]{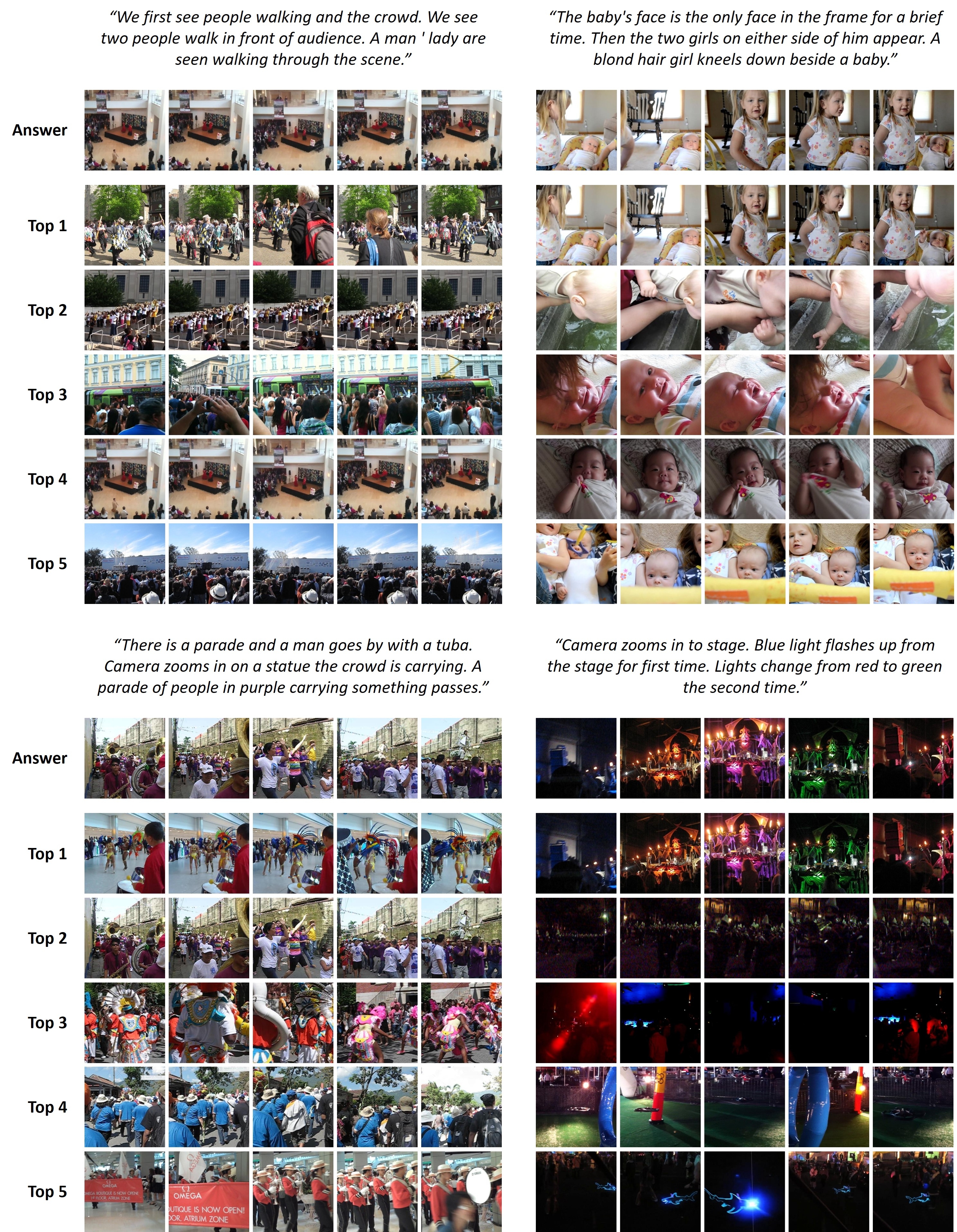}
    \caption{Qualitative examples of \textbf{zero-shot text-to-video retrieval} on DiDeMo~\cite{hendricks2017didemo}.}
    \label{fig:zs-didemo}
\end{figure*}

\begin{figure*}[t]
\centering
    \includegraphics[width=\linewidth]{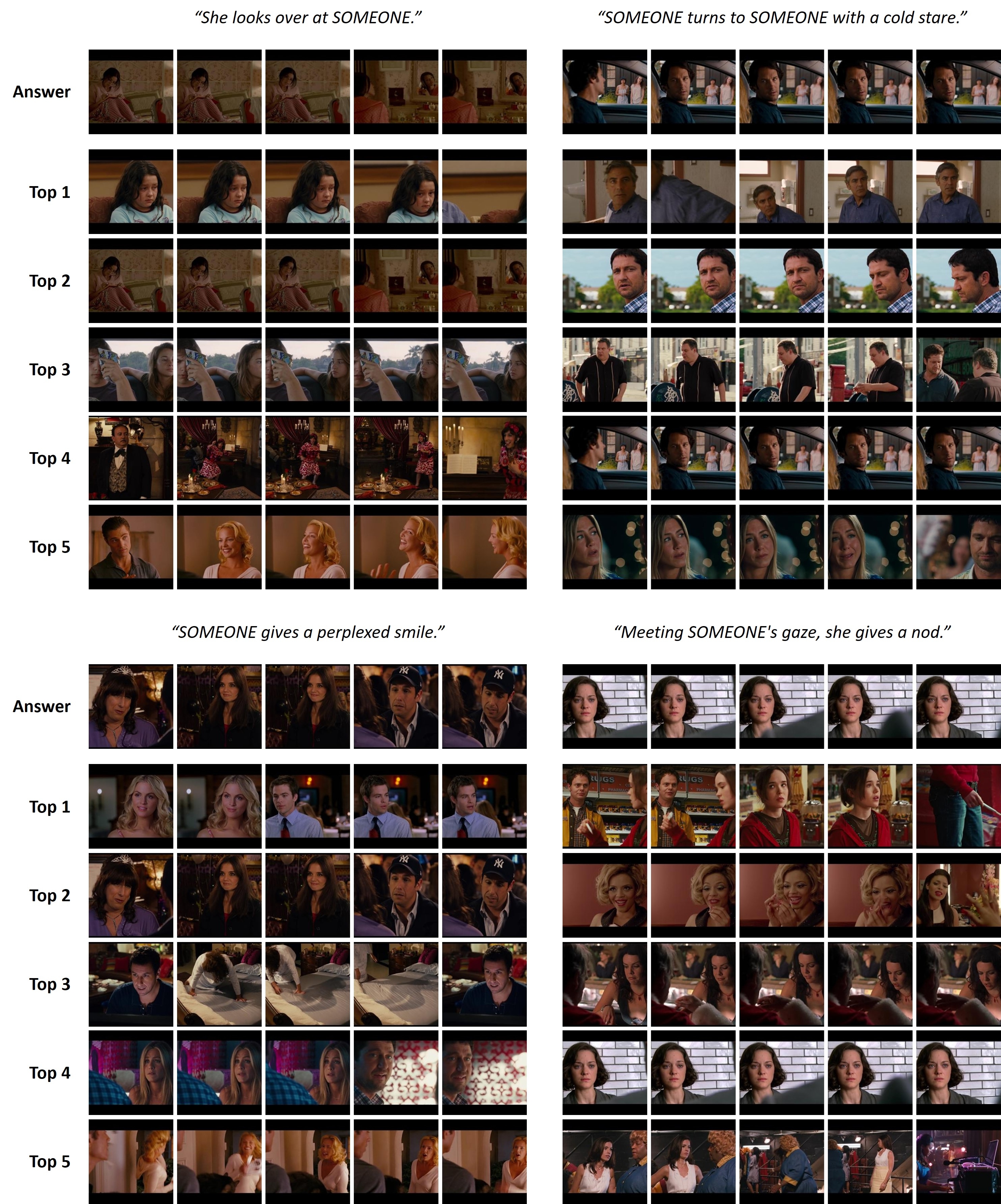}
    \caption{Qualitative examples of \textbf{zero-shot text-to-video retrieval} on LSMDC~\cite{rohrbach2015lsmdc}.}
    \label{fig:zs-lsmdc}
\end{figure*}

We also show the visualizations of zero-shot text-to-video retrieval on MSRVTT (Figure~\ref{fig:zs-msrvtt}), DiDeMo (Figure~\ref{fig:zs-didemo}), and LSMDC (Figure~\ref{fig:zs-lsmdc}) to demonstrate that MVM can help video understanding from different domains, such as gaming, animation, human activity, or movie scene.

\section{Additional Pre-training Details}
%\linjie{@ray, maybe copy the relevant paragraphs from CVPR main text?}
\noindent \textbf{Vidoe-Text Matching (VTM).}
VTM enhances the cross-modal fusion via modeling the alignments between visual and textual inputs. At each training step, we randomly replace the corresponding text $\mathcal{X}_\text{pos}$ for a given video $\mathcal{V}$ with the text description $\mathcal{X}_\text{neg}$ from a different video in the same batch. Both the positive pair $(\mathcal{V}, \mathcal{X}_\text{pos})$ and negative pair $(\mathcal{V}, \mathcal{X}_\text{neg})$ are modeled by Cross-modal Transformer (CT), and VTM is to tell them apart from the global VidL representation $h^\text{c}$ of the \texttt{[CLS]} token. In particular, $h^\text{c}$ will be processed by a fully-connected layer ($\text{FC}^\text{VTM}$) to learn contrastively through classification:
\vspace{4pt}
\begin{equation}
\begin{split}
    b^\text{pos} &= \text{FC}^\text{VTM}(h^\text{c}_\text{pos}), b^\text{neg} = \text{FC}^\text{VTM}(h^\text{c}_\text{neg}), \\
    \mathcal{L}_\text{VTM} &= - \frac{1}{B} \sum^B_i \log \frac{b^\text{pos}_i}{b^\text{pos}_i+\sum b^\text{neg}_i},
\end{split}
\end{equation}
where $h^\text{c}_\text{pos}$ or $h^\text{c}_\text{neg}$ is $h^\text{c}$ of positive or negative pairs. 

\vspace{0.5ex}
\noindent \textbf{Masked Language Modeling (MLM).}
In MLM, we randomly mask out some word tokens with a probability of 15\%.\footnote{Following BERT~\cite{devlin2019bert}, We replace 80\% of masked word tokens as the \texttt{[MASK]} token, 10\% as a random token, and 10\% as its original token.} The goal is to recover these masked word tokens $x$ from the joint VidL features $h$ modeled by CT. Specifically, the corresponding $h^\text{x}$ for these masked tokens are fed in a fully-connected layer ($\text{FC}^\text{MLM}$) and projected to the discrete token space for classification:
\vspace{4pt}
\begin{equation}
\begin{split}
    x'_i &= \text{FC}_\text{MLM}(h^\text{x}_i), \\
    \mathcal{L}_\text{MLM} &= - \mathbb{E}~[\frac{1}{|\mathcal{M}^\text{MLM}|} \sum\nolimits_{i \in \mathcal{M}^\text{MLM}} \log P(x_i~|~x'_i)],
\end{split}
\end{equation}
where $\mathcal{M}^\text{MLM}$ denotes the index set of masked word tokens.

\vspace{0.5ex}
\noindent \textbf{Implementation Details.~}
Our \modelname is implemented based on PyTorch~\cite{paszke2019pytorch}. As discussed in the main text and supported by the additional experimental results above, our final pre-training setting is ($i$) VTM+MLM+MVM (with MVM target as spatial-focused image features from Swin-B~\cite{liu2021swin},  applied on video-text inputs only) as the pre-training tasks; ($ii$) 2-layer MLP as the MVM prediction head and $l_1$ regression as the MVM loss; and ($iii$) blockwise masking + random masking with masking ratio of 30\% as the masking strategy. We adopt AdamW~\cite{loshchilov2019adamw} as the optimizer with a warmup learning rate schedule of 5e-5 peak learning rate, betas of (0.9, 0.98), and weight decay of 1e-3 for all pre-training experiments. We pre-train our model on 32 NVIDIA V100 GPUs with a batch size of 28 per GPU.  Pre-training with 10 epochs on WebVid2.5M~\cite{bain2021frozen} + CC3M~\cite{sharma2018cc} takes about 27 hours to finish. We present the training settings for all finetuning experiments in the next section.

\section{Experimental Setup of Downstream Tasks}
We test our pre-trained models on 3 popular VidL tasks across 13 downstream datasets, including text-to-video retrieval, video question answering, and video captioning. For text-to-video retrieval, we report downstream performance on MSRVTT~\cite{xu2016msrvtt}, DiDeMo~\cite{hendricks2017didemo}, and LSMDC~\cite{rohrbach2015lsmdc} and use Recall at K (R@K, K=1,5,10) as the evaluation metric. For video question answering, we consider datasets in both multiple-choice and open-ended settings, including TGIF-Action, TGIF-Transition, TGIF-Frame~\cite{jang2017tgif-qa}, MSRVTT-MC~\cite{yu2018js-fusion}, MSRVTT-QA, MSVD-QA~\cite{xu2017msrvtt-qa}, LSMDC-MC and LSMDC-FiB~\cite{torabi2016lsmdc-fib}. We evaluate our models using accuracy. For video captioning, we report CIDER scores on MSRVTT and MSVD. 

We follow the standard training/validation/testing splits of the original datasets. If not otherwise stated, we sparsely sample $T$ = 5  video frames and adopt video frame size 224 with patch size $H$ = $W$ = 32. Similar to pre-training, we use AdamW~\cite{loshchilov2019adamw} to fine-tune our model for each downstream task with a warmup learning rate schedule of 2e-5 peak learning rate, betas of (0.9, 0.98), and weight decay of 1e-3. All finetuning experiments are conducted on Microsoft Azure~\cite{msft-azure} adopting mixed-precision training with DeepSpeed~\cite{rasley2020deepspeed}.\footnote{We conduct retrieval finetuning on 8 80GB A100 GPUs to enable larger batch size, while all other finetuning experiments are conducted on 8 32GB V100 GPUs.} All video data are pre-processed by evenly extracting 32 frames to avoid expensive decoding on-the-fly. During training, we randomly sample $T$ frames from 32 frames, resize the shorter side of all frames to 224, and random crop (224x224) at the same location for all the frames in a given video. During inference, we evenly sample $T$ frames from 32 frames and center crop (224x224) for all the sampled video frames. 

\subsection{Text-To-Video Retrieval}
For text-to-video retrieval, similar to visual-text matching (VTM) during pre-training, we treat corresponding video-text pairs in the same batch as positives and all other pairwise combinations as negatives. We adopt a fully-connected (FC) layer (FC$^\text{T2V}$) over the VidL representation $h^\text{c}$ of the \texttt{[CLS]} token to learn through classification:
\vspace{4pt}
\begin{equation}
\begin{split}
    b^\text{pos}&= \text{FC}^\text{T2V}(h^\text{c}_\text{pos}), b^\text{neg}= \text{FC}^\text{T2V}(h^\text{c}_\text{neg}), \\
    \mathcal{L}_\text{T2V} &= - \frac{1}{B} \sum^{B}_{i} \log \frac{b^\text{pos}_i}{b^\text{pos}_i+\sum b^\text{neg}_i}, \\
\end{split}
\end{equation}
%\linjie{@ray, modify this to reflect the change you made for VTM in the code}
where $h^\text{c}_\text{pos}$ or $h^\text{c}_\text{neg}$ is $h^\text{c}$ of positive or negative pairs. In particular, we use pre-trained FC$^\text{VTM}$ for zero-shot text-to-video retrieval and to  initialize FC$^\text{T2V}$ for further fine-tuning on each downstream text-to-video retrieval task.

\vspace{0.5ex}
\noindent \textbf{MSRVTT~\cite{xu2016msrvtt}}
contains 10K YouTube videos with 200K human annotations. For fair comparison~\cite{bain2021frozen,lei2021clip-bert}, we train on 9K training+validation splits and evaluate on the 1K-A testing split. We adopt batch size 20 per GPU and train for 10 epochs. 

\vspace{0.5ex}
\noindent \textbf{DiDeMo~\cite{hendricks2017didemo}}
consists of 10K videos annotated with 40K sentences from Flickr. Following~\cite{bain2021frozen,lei2021clip-bert}, we concatenate all sentences from the same video into a paragraph and perform paragraph-to-video retrieval for DiDeMo. We adopt batch size 16 per GPU and train for 10 epochs. 

\vspace{0.5ex}
\noindent \textbf{LSMDC~\cite{rohrbach2015lsmdc}}
contains 118K video clips from 202 movies. Each clip has a caption from movie scripts or descriptive video services. Following~\cite{bain2021frozen,miech2019howto100m}, we evaluate on 1K testing clips that disjoint from the training+validation splits. We adopt batch size 20 per GPU and train for 5 epochs.  

\begin{table}[t]
\centering
    \tablestyle{15pt}{1.1} 
    \def \w{20pt}
    \begin{tabular}{ccc}
        \toprule
        VideoQA & Task & \#Option \\
        \midrule
        \multirow{4}{*}{\shortstack{Multiple-Choice}} & TGIF-Action~\cite{jang2017tgif-qa} & 5 \\
        ~ & TGIF-Transition~\cite{jang2017tgif-qa} & 5 \\
        ~ & MSRVTT-MC~\cite{yu2018js-fusion} & 5 \\
        ~ & LSMDC-MC~\cite{torabi2016lsmdc-fib} & 5 \\
        \midrule
        \multirow{4}{*}{\shortstack{Open-Ended}} & TGIF-Frame~\cite{jang2017tgif-qa} & - \\
        ~ & MSRVTT-QA~\cite{xu2017msrvtt-qa} & - \\
        ~ & MSVD-QA~\cite{chen2011msvd-qa} & - \\
        ~ & LSMDC-FiB~\cite{torabi2016lsmdc-fib} & - \\
        \bottomrule
    \end{tabular}
    \caption{Summary of \textbf{video question answering} tasks. For open-ended Video QA, we do not limit the answer vocabulary to a fixed candidate set.}
    \label{table:vqa}
\end{table}

\subsection{Video Question Answering}
We test our model on video question answering (QA) tasks in both multiple-choice and open-ended settings as Table~\ref{table:vqa}. We follow LAVENDER~\cite{li2022lavender} to formulate Video QA as Masked Language Modeling due to its superior performance. For multiple-choice QA tasks, we concatenate question with all answer options and add a \texttt{[MASK]} to form the input text (\texttt{Q+A0+A1+A2+A3+A4+[MASK]}). We treat the same Masked Language Modeling (MLM) layer as used in pre-training upon $h^\text{x}$ to predict the word token corresponding to the answer index (\textit{e.g.}, \texttt{0,1,2,3,4}). Similarly, for open-ended QA tasks, we apply MLM over the input (\texttt{Q+[MASK]}). Cross-entropy loss is used to supervise the downstream finetuning over the whole word vocabulary.

\vspace{0.5ex}
\noindent \textbf{TGIF-Action, TGIF-Transition, and TGIF-Frame~\cite{jang2017tgif-qa}}
require spatial-temporal reasoning to answer questions regarding GIF videos in TGIF-QA Specifically, we aim to test our model along three dimensions: ($i$) \textbf{Action}: to recognize the repeated action; ($ii$) \textbf{Transition}: to identify the transition between the before and after states; ($iii$) \textbf{Frame}: to answer questions about a specific frame from the GIF video. Among them, TGIF-Action and TGIF-Transition are collected under a multiple-choice setting, and TGIF-Frame is an open-ended video QA task with free-form answers. We adopt batch size 24 and train for 56/20/10 epochs for Action/Transition/Frame, respectively.

\vspace{0.5ex}
\noindent \textbf{MSRVTT-MC~\cite{yu2018js-fusion} and MSRVTT-QA~\cite{xu2017msrvtt-qa}}
are created based on videos and captions in MSRVTT~\cite{xu2016msrvtt}. MSRVTT-MC is a multiple-choice task with videos as questions, and captions as answers. Each video contains 5 captions, with only one positive match. This setting can be viewed as video-to-text retrieval, hence we simply evaluate the model trained on MSRVTT-Retrieval.  MSRVTT-QA contains 243K open-ended questions over 10K videos. We adopt batch size 24 per GPU and training epochs 8.

\vspace{0.5ex}
\noindent \textbf{MSVD-QA~\cite{xu2017msrvtt-qa}}
consists of 47K open-ended questions over 2K videos, based on video-caption pairs from MSVD~\cite{chen2011msvd-qa}. We adopt batch size 24 per GPU and train for 10 epochs.

\vspace{0.5ex}
\noindent \textbf{LSMDC-MC and LSMDC-FiB~\cite{torabi2016lsmdc-fib}} are built from the LSMDC dataset~\cite{rohrbach2015lsmdc}. Similar to MSRVTT-MC, LSMDC-MC requires the model to select the only positive caption that describes the video from 5 caption candidates, and can be formulated as video-to-text retrieval. LSMDC-FiB replaces a word in the question sentence with the \texttt{[BLANK]} token, and requires the model to recover the missing word.  We regard LSMDC-FiB as an open-ended Video QA task. In particular, we replace the \texttt{[BLANK]} token with \texttt{[MASK]} token, and use the MLM prediction head over the representation $h_x$ of the \texttt{[MASK]} token to predict the correct answer. We adopt batch size 24 per GPU and train for 10 epochs.

\subsection{Video Captioning}
For video captioning, we evaluate on MSRVTT~\cite{xu2016msrvtt} and MSVD~\cite{chen2011msvd}. \textbf{MSRVTT} consists of 10K videos with 20 captions per video, and \textbf{MSVD} contains 2K videos, with 40 captions per video. We follow the standard captioning splits to train/evaluate with \modelname. The captioning finetuning is formulated as masked language modeling (MLM) with a causal attention mask so that the current word token only attends to the tokens before it, following SwinBERT~\cite{lin2022swin-bert}. During training, we set the probability of random masking caption tokens to 0.15, the same as what is used in MLM during pre-training. We adopt batch size 24 per GPU and train for 20 epochs. During inference, we generate the captions auto-regressively. At each generation step, a \texttt{[MASK]} token is appended to the previously generated tokens, and the model will predict the current tokens based on the learned embedding at the \texttt{[MASK]} token position. We perform generation until the model outputs a \texttt{[SEP]}, which is defined as the sentence ending token or when it reaches the maximum generation step 50. 
\clearpage
\clearpage
\clearpage

%% file: tabs/violet_v1_vs_v2.tex
\vspace{-2ex}
\begin{table}[H]
\centering
    \tablestyle{5pt}{1.2} 
    \def \w{20pt} 
    % \resizebox{\linewidth}{!}{
    \begin{tabular}{lc|ccccc}
        \shline
        Method & MVM & TGIF-Frame & \multicolumn{4}{c}{DiDeMo-Retrieval} \\
        \cmidrule(lr){3-3}\cmidrule(lr){4-7}
         & Target & Acc. & R1 & R5 & R10 & AveR \\
        \hline
        VIOLET~\cite{fu2021violet} & VQ & 70.5 &  32.9 & 63.0 & 74.5 & 56.8\\
         VIOLETv2 & SIF &  \textbf{72.8} & \textbf{47.9} & \textbf{76.5} & \textbf{84.1} & \textbf{69.5}\\
        \shline
    \end{tabular}
    % }
    \caption{\textbf{Fair comparison to \modelorig~\cite{fu2021violet}.} Both
models are pre-trained on WebVid~\cite{bain2021frozen}+CC~\cite{sharma2018cc}.}
    \label{table:violet-v1-v2}
\end{table}

%% file: tabs/supp_fom.tex
\vspace{-2ex}
\begin{table}[H]
\centering
    \tablestyle{7pt}{1.2} 
    \def \w{20pt} 
    % \resizebox{\linewidth}{!}{
    \begin{tabular}{l| ccccc}
        \shline
        \multirow{2}{*}{VTM+MLM+} & TGIF-Frame & \multicolumn{4}{c}{DiDeMo-Retrieval} \\
        \cmidrule(lr){2-2}\cmidrule(lr){3-6}
         & Acc. & R1 & R5 & R10 & AveR \\
        \hline
         MVM (SIF)  & \textbf{68.8} & \textbf{35.1} & \textbf{63.3} & \textbf{73.1} & \textbf{57.2} \\
         FOM  &  68.1 & 27.6 & 59.0 &  70.7 & 52.4\\
        \shline
    \end{tabular}
    % }
    \caption{\textbf{MVM vs. FOM}. Both
models are pre-trained on WebVid~\cite{bain2021frozen}.}
    \label{table:fom}
\end{table}

%% file: tabs/supp_vid_init.tex
\vspace{-2ex}
\begin{table}[H]
\centering
    \tablestyle{3.5pt}{1.2} 
    \def \w{20pt} 
    % \resizebox{\linewidth}{!}{
    \begin{tabular}{lc| ccccc}
        \shline
        \multirow{2}{*}{Weight Init.} & MVM & TGIF-Frame & \multicolumn{4}{c}{DiDeMo-Retrieval} \\
        \cmidrule(lr){3-3}\cmidrule(lr){4-7}
        & on VL & Acc. & R1 & R5 & R10 & AveR \\
        \hline
         % Random & \ding{53} & 55.9 & 5.6 & 19.9 & 29.8 & 18.5\\
         % &  \checkmark &  56.5 & 7.4 & 22.9 & 33.8 & 21.4\\
         % %\hline
         % %Swin~\cite{liu2021swin} &\ding{53} & 68.0 & 31.7 & 61.7 & 72.0 & 55.1\\
         % %&  \checkmark & 68.0 & 31.7 & 60.7 & 70.7 %& 54.3\\
         % \hline
         % VidSwin-B~\cite{liu2022video-swin}) &\ding{53} & 68.1 & 28.7 & 57.0 & 69.7 & 51.8 \\
         % \rowcolor{lightgray}
         %  & \checkmark & \textbf{68.8} & \textbf{35.1} & \textbf{63.3} & \textbf{73.1} & \textbf{57.2} \\
          Random & \ding{53} & 55.9 & 5.6 & 19.9 & 29.8 & 18.5 \\
        &  \checkmark &  56.5 & 7.4 & 22.9 & 33.8  & 21.4\\
        %&  \checkmark & 68.0 & 31.7 & 60.7 & 70.7 %& 54.3\\
        \hline
        V-only MVM (TVF) & \ding{53} & 59.9 & 15.3 & 38.4 & 54.7 &36.1\\
        % {'r@1': '15.32', 'r@5': '38.41', 'r@10': '54.65', 'median': '9'}
        & \checkmark & 60.2  & 17.4 & 43.2  & 56.0 & 38.9\\
        \hline
        V-only MVM (SIF) & \ding{53} & 61.0 & 16.9 & 42.4 & 54.9 & 38.1 \\
        & \checkmark & 61.5  & 18.6 & 44.0 & 58.1 & 40.2\\
        %\hline
        %Swin~\cite{liu2021swin} &\ding{53} & 68.0 & 31.7 & 61.7 & 72.0 & 55.1\\
        %&  \checkmark & 68.0 & 31.7 & 60.7 & 70.7 %& 54.3\\
        \hline
        VidSwin-B &\ding{53} & 68.1 & 28.7 & 57.0 & 69.7 & 51.8\\
        \rowcolor{lightgray} & \checkmark & \textbf{68.8} & \textbf{35.1} & \textbf{63.3} & \textbf{73.1} & \textbf{57.2}\\
        \shline
    \end{tabular}
    % }
    \caption{Impact of \textbf{weight initialization and learning of video backbone}. All variants are pre-trained on video-text from WebVid~\cite{bain2021frozen} for 5 epochs. The MVM target is spatial-focused image features (SIF) from Swin-B~\cite{liu2021swin}), if not specified otherwise. For V-only MVM (TVF/SVF), we first self-supervisedly pre-train the video backbone with MVM on video-only inputs from WebVid for 5 epochs. The final pre-training setting is highlighted in \colorbox{lightgray}{gray}.}
    \label{table:mvm-webvid-init}
\end{table}

%% file: tabs/supp_mvm_loss.tex
\vspace{-2ex}
\begin{table}[H]
\centering
    \tablestyle{8pt}{1.2} 
    \def \w{20pt} 
    % \resizebox{\linewidth}{!}{
    \begin{tabular}{c| ccccc}
        \shline
        \multirow{2}{*}{MVM Loss} & TGIF-Frame & \multicolumn{4}{c}{DiDeMo-Retrieval} \\
        \cmidrule(lr){2-2}\cmidrule(lr){3-6}
        & Acc. & R1 & R5 & R10 & AveR \\
        \hline
        \rowcolor{lightgray}
         $l_1$ & \textbf{\siftgif} & \textbf{\sifdro} & \textbf{\sifdrf} & \textbf{\sifdrt} & \textbf{\sifaver}\\
         $l_2$ & \textbf{68.8} & 33.0 & 60.1 & 71.9 & 55.0\\
         % {'r@1': '32.99', 'r@5': '60.06', 'r@10': '71.91', 'median': '3'}
        \shline
    \end{tabular}
    % }
    \caption{Impact of \textbf{MVM loss type}. All variants are pre-trained on WebVid~\cite{bain2021frozen} with VTM+MLM+MVM (SIF) for 5 epochs, using RM as the masking strategy with ratio of 15\%. The final pre-training setting is highlighted in \colorbox{lightgray}{gray}.}
    \label{table:mvm-webvid-loss}
\end{table}

%% file: tabs/supp_mvm_head.tex
\vspace{-2ex}
\begin{table}[H]
\centering
    \tablestyle{7pt}{1.2} 
    \def \w{20pt} 
    % \resizebox{\linewidth}{!}{
    \begin{tabular}{c| ccccc}
        \shline
        \multirow{2}{*}{MVM Head} & TGIF-Frame & \multicolumn{4}{c}{DiDeMo-Retrieval} \\
        \cmidrule(lr){2-2}\cmidrule(lr){3-6}
         & Acc. & R1 & R5 & R10 & AveR \\
        \hline
         1 Linear Layer & \textbf{68.8} & 31.3 & 60.1 & 72.8 & 54.7\\
         \rowcolor{lightgray}
         2-layer MLP & \textbf{\siftgif} & \textbf{\sifdro} & \textbf{\sifdrf} & \textbf{\sifdrt} & \textbf{\sifaver}\\ 
        \shline
    \end{tabular}
    % }
    \caption{Impact of \textbf{MVM prediction head}. All
variants are pre-trained on WebVid~\cite{bain2021frozen} with VTM+MLM+MVM (SIF) for 5 epochs, using RM as the masking strategy with ratio of 15\%. The final pre-training setting is highlighted in \colorbox{lightgray}{gray}.}
    \label{table:mvm-webvid-head}
    \vspace{-0.5ex}
\end{table}

%% file: tabs/supp_tvf.tex
\vspace{-2ex}
\begin{table}[H]
\centering
    \tablestyle{4.5pt}{1.2} 
    \def \w{20pt} 
    % \resizebox{\linewidth}{!}{
    \begin{tabular}{c| ccccc}
        \shline
        \multirow{2}{*}{MVM Target} & TGIF-Frame & \multicolumn{4}{c}{DiDeMo-Retrieval} \\
        \cmidrule(lr){2-2}\cmidrule(lr){3-6}
        & Acc. & R1 & R5 & R10 & AveR \\
        \hline
         TVF (VidSwin-L~\cite{liu2022video-swin}) & 68.0 & 32.8 & 60.5 & 73.0 & 55.4\\
         TVF (VidSwin-B) & 67.5 & 25.8 & 55.0 & 68.0 & 49.6\\
         \hline
         TVF-dense (VidSwin-L) & \textbf{68.4} & \textbf{34.3} & \textbf{60.8} & \textbf{72.4} & \textbf{55.8}\\
         % & Temporal-aware Video Features (VidSwin-B~\cite{liu2022ideo-swin}) & 67.5 {\color{BrickRed}(-0.6)} & 25.8 {\color{BrickRed}(-2.9)} & 55.0 {\color{BrickRed}(-2.0)} & 68.0 {\color{BrickRed}(-1.7)} & 49.6 {\color{BrickRed}(-2.2)} \\
        \shline
    \end{tabular}
    % }
    \caption{\textbf{Temporal-aware video feature (TVF) target models \textit{vs.} downstream performance}. All variants are pre-trained on WebVid~\cite{bain2021frozen} with VTM+MLM+MVM (TVF) for 5 epochs, using RM as the masking strategy with ratio of 15\%. }
    \label{table:tvf}
\end{table}

%% file: tabs/supp_combine_target.tex
\vspace{-4ex}
\begin{table}[H]
\centering
    \tablestyle{7pt}{1.2} 
    \def \w{20pt} 
    % \resizebox{\linewidth}{!}{
    \begin{tabular}{l| ccccc}
        \shline
        \multirow{2}{*}{MVM Targets} & TGIF-Frame & \multicolumn{4}{c}{DiDeMo-Retrieval} \\
        \cmidrule(lr){2-2}\cmidrule(lr){3-6}
         & Acc. & R1 & R5 & R10 & AveR \\
        \hline
         \rowcolor{lightgray}
          SIF & \siftgif & \textbf{\sifdro} & \sifdrf & \textbf{\sifdrt} & \textbf{\sifaver}\\
         TVF & 68.0 & 32.8 & 60.5 & 73.0 & 55.4\\
        \hline
        SIF + TVF & \textbf{69.2} & 33.8 & \textbf{63.0} & 74.4 & 57.1\\
        \shline
    \end{tabular}
    % }
    \caption{\textbf{Combining target features for MVM}. All
variants are pre-trained on WebVid~\cite{bain2021frozen} for 5 epochs. The final pre-training setting is highlighted in \colorbox{lightgray}{gray}.}
    \label{table:mvm-webvid-combine-sif-tvf}
\end{table}

%% file: tabs/supp_deit.tex
\begin{table}[t]
 \centering 
 \tablestyle{4.5pt}{1.2}
    % \resizebox{\columnwidth}{!}{%
    \begin{tabular}{l cc|cccc}
        \shline
        Image Feat. & Train & IN-1K & TGIF-Frame & \multicolumn{3}{c}{DiDeMo-Retrieval} \\
        \cmidrule(lr){3-3} \cmidrule(lr){4-4} \cmidrule(lr){5-7}
         Model & Data & ACC@1 & Acc. & R1 & R5 & R10  \\
        \hline
        ResNet-50 & IN-1K & 76.1 & 67.3 & 29.1 & 58.1 & 69.3 \\
        DeiT~\cite{touvron2021training} & IN-1K & 83.4 & \textbf{68.4} & 31.4 & 59.4 & 72.2 \\
        Swin-B & IN-1K & \textbf{83.5} & 68.3 & \textbf{34.9} & \textbf{63.4} & \textbf{73.9} \\
        \shline
    \end{tabular}
    % }
    \caption{\textbf{Comparing Swin-B vs. another SIF model (DEiT)} All variants are pre-trained on WebVid with VTM+MLM+MVM (SIF) for 5 epochs, using RM with 15\% as the masking strategy.}
    % \caption{\small \linjie{DEIT, ls 0.02 69.2, ls 0.04 68.9, ls 1, 67.2, retrieval results from ls 0.02}}
    \label{table:mvm-sif-deit}
\end{table}

%% file: tabs/supp_clip.tex
\vspace{-4ex}
\begin{table}[H]
\centering \tablestyle{5pt}{1.2}
    %\def \w{20pt}
    % \resizebox{\linewidth}{!}{
    \begin{tabular}{ll| cccc}
        \shline
        MVM & \multirow{2}{*}{Settings}  & TGIF-Frame & \multicolumn{3}{c}{DiDeMo-Retrieval} \\
        \cmidrule(lr){3-3}\cmidrule(lr){4-6}
        target & & Acc. & R1 & R5 & R10 \\
        \hline
        CLIP & \multirow{2}{*}{Default} & 67.7 & 29.8 & 57.8 & 68.5 \\
        Swin-B &  & 68.8 & 35.1 & 63.3 & 73.1 \\
        \hline
        \multirow{5}{*}{CLIP} & lr $\times$ 2 &70.5  & 32.9 & 61.6 & 73.5 \\
        % & lr / 2 (1e-5) & 67.0 & 26.1 & 54.8 & 67.8 \\
        & masking ratio = 0.3 & 68.0 & 31.8 & 59.6 & 71.3 \\
        & loss type = $l_2$ & 68.3 & 30.1 & 59.1 & 71.0 \\
        & linear MVM head & 68.2 & 30.5 & 58.3 & 69.2\\
        \hline
       \multirow{2}{*}{Swin-B} & lr $\times$ 2 & \textbf{70.6} & 33.3 & 63.7 & \textbf{75.2}\\
       & masking ratio = 0.3  & 68.8 & \textbf{36.2} & \textbf{64.0} & 74.5 \\
        \shline
    \end{tabular}
    \caption{\textbf{Investigation of Training Recipe with CLIP Target}. All variants are pre-trained on WebVid with VTM+MLM+MVM for 5 epochs. The default setting follows Table 1 in the main text, that is RM with 15\% masking ratio, $l_1$ loss and 2-layer MLP head for MVM prediction.}
    \label{table:mvm-clip}
\end{table}

%% file: tabs/supp_extended_table1.tex
\begin{table*}[t]
\centering
    \tablestyle{6pt}{1.1} 
    \def \w{20pt} 
    % \resizebox{\linewidth}{!}{
    \begin{tabular}{ll| ccccc}
        \shline
        \multirow{2}{*}{Pre-training Tasks} & \multirow{2}{*}{MVM Target} & MSVD-QA & \multicolumn{4}{c}{MSRVTT-Retrieval} \\
        \cmidrule(lr){3-3}\cmidrule(lr){4-7}
         &  & Acc. & R1 & R5 & R10 & AveR \\
        \hline
        % None & None &  \\
        VTM+MLM & None &  49.2  &  26.0 & 56.6 & 69.4 & 50.7\\
        \hline
        \multirow{7}{*}{+MVM} & RGB Pixel Values & 51.0 {\color{ForestGreen}(+1.8)}  & 27.4 {\color{ForestGreen}(+1.4)} & 58.0 {\color{ForestGreen}(+1.4)}& 69.8 {\color{ForestGreen}(+0.4)} & 51.7 {\color{ForestGreen}(+1.0)}\\
          & Histogram of Oriented Gradients~\cite{dalal2005hog} & 50.1 {\color{ForestGreen}(+0.9)} & 27.4 {\color{ForestGreen}(+1.4)} & 57.7 {\color{ForestGreen}(+1.1)}& 70.2 {\color{ForestGreen}(+0.8)} & 51.8 {\color{ForestGreen}(+1.1)}\\
          \cline{2-7} 
        & Depth Maps (DPT-L~\cite{ranftl2021dpt}) & 50.3 {\color{ForestGreen}(+1.1)}  & 28.0 {\color{ForestGreen}(+2.0)} & 57.4 {\color{ForestGreen}(+0.8)}& 70.6 {\color{ForestGreen}(+0.8)} & 52.0 {\color{ForestGreen}(+1.3)}\\
         & Optical Flow (RAFT-L~\cite{teed2020raft}) & 49.7 {\color{ForestGreen}(+0.5)}  & 25.8 {\color{BrickRed}(-0.2)} & 55.8 {\color{BrickRed}(-0.8)}& 69.4  & 50.3 {\color{BrickRed}(-0.4)}\\
         \cline{2-7} 
         \rowcolor{lightgray}
          & Spatial-focused Image Features (Swin-B~\cite{liu2021swin}) & \textbf{51.1 {\color{ForestGreen}(+1.9)}} &  29.4 \space {\color{ForestGreen}(+3.4)} & \textbf{59.9 \space {\color{ForestGreen} (+3.6)}} & \textbf{73.1 \space {\color{ForestGreen}(+3.7)}} & \textbf{54.1 \space {\color{ForestGreen}(+3.4)}} \\
          % & Temporal-aware Video Features (VidSwin-B~\cite{liu2022ideo-swin}) & 67.5 {\color{BrickRed}(-0.6)} & 25.8 {\color{BrickRed}(-2.9)} & 55.0 {\color{BrickRed}(-2.0)} & 68.0 {\color{BrickRed}(-1.7)} & 49.6 {\color{BrickRed}(-2.2)} \\
          & Temporal-aware Video Features (VidSwin-L~\cite{liu2022video-swin}) & 49.8 {\color{ForestGreen}(+0.6)} &  29.9 {\color{ForestGreen}(+3.9)} &  58.1 {\color{ForestGreen}(+1.5)} &  70.2 {\color{ForestGreen}(+0.8)} &  52.7 {\color{ForestGreen}(+2.0)} \\
          % {'r@1': '32.79', 'r@5': '60.47', 'r@10': '73.03', 'median': '3'}
          \cline{2-7}
         & Discrete Visual Tokens (DALL-E~\cite{ramesh2021dalle}) & 50.7 {\color{ForestGreen}(+1.5)} & 27.3 {\color{ForestGreen}(+1.3)} & 58.3 {\color{ForestGreen}(+1.7)} & 70.0 {\color{ForestGreen}(+0.6)} & 51.9 {\color{ForestGreen}(+1.2)} \\
          & Multimodal Features (CLIP-ViT-B~\cite{radford2021clip}) & 50.2 {\color{ForestGreen}(+1.0)} & \textbf{30.0 {\color{ForestGreen}(+4.0)}} & 58.8 {\color{ForestGreen}(+2.2)}& 71.1 {\color{ForestGreen}(+1.7)} & 53.3 {\color{ForestGreen}(+2.6)}\\
        \shline
    \end{tabular}
    % }
    \caption{\textbf{Comparing target features for MVM applied to video-text data}. All variants are pre-trained on WebVid~\cite{bain2021frozen} for 5 epochs. Masking is performed randomly (RM) with a ratio of 15\%. The final pre-training setting is highlighted in \colorbox{lightgray}{gray}. }
    \label{table:mvm-webvid-supp}
    \vspace{1ex}
\end{table*}

%% file: arxiv.bbl
\begin{thebibliography}{10}\itemsep=-1pt

\bibitem{msft-azure}
{Microsoft Azure}.
\newblock \url{https://azure.microsoft.com/}.

\bibitem{anderson2018bottom}
Peter Anderson, Xiaodong He, Chris Buehler, Damien Teney, Mark Johnson, Stephen
  Gould, and Lei Zhang.
\newblock {Bottom-Up and Top-Down Attention for Image Captioning and Visual
  Question Answering}.
\newblock In {\em Conference on Computer Vision and Pattern Recognition
  (CVPR)}, 2018.

\bibitem{bain2021frozen}
Max Bain, Arsha Nagrani, Gül Varol, and Andrew Zisserman.
\newblock {Frozen in Time: A Joint Video and Image Encoder for End-to-End
  Retrieval}.
\newblock In {\em International Conference on Computer Vision (ICCV)}, 2021.

\bibitem{bao2022beit}
Hangbo Bao, Li Dong, and Furu Wei.
\newblock {BEiT: BERT Pre-Training of Image Transformers}.
\newblock In {\em International Conference for Learning Representations
  (ICLR)}, 2022.

\bibitem{buch2022revisiting}
Shyamal Buch, Cristóbal Eyzaguirre, Adrien Gaidon, Jiajun Wu, Li Fei-Fei, and
  Juan~Carlos Niebles.
\newblock {Revisiting the "Video" in Video-Language Understanding}.
\newblock In {\em Conference on Computer Vision and Pattern Recognition
  (CVPR)}, 2022.

\bibitem{carreira2017quo}
Joao Carreira and Andrew Zisserman.
\newblock {Quo Vadis, Action Recognition? A New Model and the Kinetics
  Dataset}.
\newblock In {\em Conference on Computer Vision and Pattern Recognition
  (CVPR)}, 2017.

\bibitem{chen2011msvd}
David~L. Chen and William~B. Dolan.
\newblock {Collecting Highly Parallel Data for Paraphrase Evaluation}.
\newblock In {\em ACL}, 2011.

\bibitem{chen2011msvd-qa}
David~L. Chen and William~B. Dolan.
\newblock {Collecting Highly Parallel Data for Paraphrase Evaluation}.
\newblock In {\em Annual Meetings of the Association for Computational
  Linguistics (ACL)}, 2011.

\bibitem{chen2020uniter}
Yen-Chun Chen, Linjie Li, Licheng Yu, Ahmed~El Kholy, Faisal Ahmed, Zhe Gan, Yu
  Cheng, and Jingjing Liu.
\newblock {UNITER: UNiversal Image-TExt Representation Learning}.
\newblock In {\em European Conference on Computer Vision (ECCV)}, 2020.

\bibitem{dalal2005hog}
Navneet Dalal and Bill Triggs.
\newblock {Histograms of Oriented Gradients for Human Detection}.
\newblock In {\em Conference on Computer Vision and Pattern Recognition
  (CVPR)}, 2005.

\bibitem{deng2009imagenet}
Jia Deng, Wei Dong, Richard Socher, Li-Jia Li, Kai Li, and Li Fei-Fei.
\newblock {ImageNet: a Large-Scale Hierarchical Image Database}.
\newblock In {\em Conference on Computer Vision and Pattern Recognition
  (CVPR)}, 2009.

\bibitem{devlin2019bert}
Jacob Devlin, Ming-Wei Chang, Kenton Lee, and Kristina Toutanova.
\newblock {BERT: Pre-training of Deep Bidirectional Transformers for Language
  Understanding}.
\newblock In {\em Conference of the North American Chapter of the Association
  for Computational Linguistics (NAACL)}, 2019.

\bibitem{dosovitskiy2021vit}
Alexey Dosovitskiy, Lucas Beyer, Alexander Kolesnikov, Dirk Weissenborn,
  Xiaohua Zhai, Thomas Unterthiner, Mostafa Dehghani, Matthias Minderer, Georg
  Heigold, Sylvain Gelly, Jakob Uszkoreit, and Neil Houlsby.
\newblock {An Image is Worth 16x16 Words: Transformers for Image Recognition at
  Scale}.
\newblock In {\em International Conference for Learning Representations
  (ICLR)}, 2021.

\bibitem{dou2022empirical}
Zi-Yi Dou, Yichong Xu, Zhe Gan, Jianfeng Wang, Shuohang Wang, Lijuan Wang,
  Chenguang Zhu, Pengchuan Zhang, Lu Yuan, Nanyun Peng, Zicheng Liu, and
  Michael Zeng.
\newblock {An Empirical Study of Training End-to-End Vision-and-Language
  Transformers}.
\newblock In {\em Conference on Computer Vision and Pattern Recognition
  (CVPR)}, 2022.

\bibitem{fan2019heterogeneous}
Chenyou Fan, Xiaofan Zhang, Shu Zhang, Wensheng Wang, Chi Zhang, and Heng
  Huang.
\newblock {Heterogeneous Memory Enhanced Multimodal Attention Model for Video
  Question Answering}.
\newblock In {\em Conference on Computer Vision and Pattern Recognition
  (CVPR)}, 2019.

\bibitem{feichtenhofer2019slowfast}
Christoph Feichtenhofer, Haoqi Fan, Jitendra Malik, and Kaiming He.
\newblock {SlowFast Networks for Video Recognition}.
\newblock In {\em Conference on Computer Vision and Pattern Recognition
  (CVPR)}, 2019.

\bibitem{fu2021violet}
Tsu-Jui Fu, Linjie Li, Zhe Gan, Kevin Lin, William~Yang Wang, Lijuan Wang, and
  Zicheng Liu.
\newblock {VIOLET: End-to-End Video-Language Transformers with Masked
  Visual-token Modeling}.
\newblock In {\em arXiv:2111.1268}, 2021.

\bibitem{gabeur2020mmt}
Valentin Gabeur, Chen Sun, Karteek Alahari, and Cordelia Schmid.
\newblock {Multi-modal Transformer for Video Retrieval}.
\newblock In {\em European Conference on Computer Vision (ECCV)}, 2020.

\bibitem{gan2022vlp}
Zhe Gan, Linjie Li, Chunyuan Li, Lijuan Wang, Zicheng Liu, and Jianfeng Gao.
\newblock {Vision-Language Pre-training: Basics, Recent Advances, and Future
  Trends}.
\newblock In {\em Foundations and Trends in Computer Graphics and Vision},
  2022.

\bibitem{gao2018motion}
Jiyang Gao, Runzhou Ge, Kan Chen, and Ram Nevatia.
\newblock {Motion-Appearance Co-Memory Networks for Video Question Answering}.
\newblock In {\em Conference on Computer Vision and Pattern Recognition
  (CVPR)}, 2018.

\bibitem{gao2017tall}
Jiyang Gao, Chen Sun, Zhenheng Yang, and Ram Nevatia.
\newblock {TALL: Temporal Activity Localization via Language Query}.
\newblock In {\em International Conference on Computer Vision (ICCV)}, 2017.

\bibitem{ge2022bridge-former}
Yuying Ge, Yixiao Ge, Xihui Liu, Dian Li, Ying Shan, Xiaohu Qie, and Ping Luo.
\newblock {BridgeFormer: Bridging Video-text Retrieval with Multiple Choice
  Questions}.
\newblock In {\em Conference on Computer Vision and Pattern Recognition
  (CVPR)}, 2022.

\bibitem{he2022mae}
Kaiming He, Xinlei Chen, Saining Xie, Yanghao Li, Piotr Dollár, and Ross
  Girshick.
\newblock {Masked Autoencoders Are Scalable Vision Learners}.
\newblock In {\em Conference on Computer Vision and Pattern Recognition
  (CVPR)}, 2022.

\bibitem{he2016resnet}
Kaiming He, Xiangyu Zhang, Shaoqing Ren, and Jian Sun.
\newblock {Deep Residual Learning for Image Recognition}.
\newblock In {\em Conference on Computer Vision and Pattern Recognition
  (CVPR)}, 2016.

\bibitem{hendricks2017local-moment}
Lisa~Anne Hendricks, Oliver Wang, Eli Shechtman, Josef Sivic, Trevor Darrell,
  and Bryan Russell.
\newblock {Localizing Moments in Video with Natural Language}.
\newblock In {\em International Conference on Computer Vision (ICCV)}, 2017.

\bibitem{hendricks2017didemo}
Lisa~Anne Hendricks, Oliver Wang, Eli Shechtman, Josef Sivic, Trevor Darrell,
  and Bryan Russell.
\newblock {Localizing Moments in Video with Natural Language}.
\newblock In {\em International Conference on Computer Vision (ICCV)}, 2017.

\bibitem{jang2017tgif-qa}
Yunseok Jang, Yale Song, Youngjae Yu, Youngjin Kim, and Gunhee Kim.
\newblock {TGIF-QA: Toward Spatio-Temporal Reasoning in Visual Question
  Answering}.
\newblock In {\em Conference on Computer Vision and Pattern Recognition
  (CVPR)}, 2017.

\bibitem{jiang2020dac}
Jianwen Jiang, Ziqiang Chen, Haojie Lin, Xibin Zhao, and Yue Gao.
\newblock {Divide and Conquer: Question-Guided Spatio-Temporal Contextual
  Attention for Video Question Answering}.
\newblock In {\em AAAI Conference on Artificial Intelligence (AAAI)}, 2020.

\bibitem{kay2017kinetics}
Will Kay, Joao Carreira, Karen Simonyan, Brian Zhang, Chloe Hillier, Sudheendra
  Vijayanarasimhan, Fabio Viola, Tim Green, Trevor Back, Paul Natsev, Mustafa
  Suleyman, and Andrew Zisserman.
\newblock {The Kinetics Human Action Video Dataset}.
\newblock In {\em arXiv:1705.06950}, 2017.

\bibitem{kim2021sspt-crl-vqa}
Seonhoon Kim, Seohyeong Jeong, Eunbyul Kim, Inho Kang, and Nojun Kwak.
\newblock {Self-supervised Pre-training and Contrastive Representation Learning
  for Multiple-choice Video QA}.
\newblock In {\em AAAI Conference on Artificial Intelligence (AAAI)}, 2021.

\bibitem{krishna2017dense-caption}
Ranjay Krishna, Kenji Hata, Frederic Ren, Li Fei-Fei, and Juan~Carlos Niebles.
\newblock {Dense-Captioning Events in Videos}.
\newblock In {\em International Conference on Computer Vision (ICCV)}, 2017.

\bibitem{krishna2017vg}
Ranjay Krishna, Yuke Zhu, Oliver Groth, Justin Johnson, Kenji Hata, Joshua
  Kravitz, Stephanie Chen, Yannis Kalantidis, Li-Jia Li, David~A. Shamma,
  Michael~S. Bernstein, and Fei-Fei Li.
\newblock {Visual Genome: Connecting Language and Vision Using Crowdsourced
  Dense Image Annotations}.
\newblock In {\em International Journal of Computer Vision (IJCV)}, 2017.

\bibitem{lan2019albert}
Zhenzhong Lan, Mingda Chen, Sebastian Goodman, Kevin Gimpel, Piyush Sharma, and
  Radu Soricut.
\newblock {ALBERT: A Lite BERT for Self-supervised Learning of Language
  Representations}.
\newblock In {\em International Conference for Learning Representations
  (ICLR)}, 2020.

\bibitem{le2020hcr-vqa}
Thao~Minh Le, Vuong Le, Svetha Venkatesh, and Truyen Tran.
\newblock {Hierarchical Conditional Relation Networks for Video Question
  Answering}.
\newblock In {\em Conference on Computer Vision and Pattern Recognition
  (CVPR)}, 2020.

\bibitem{lei2021qvhighlights}
Jie Lei, Tamara~L Berg, and Mohit Bansal.
\newblock {QVHighlights: Detecting Moments and Highlights in Videos via Natural
  Language Queries}.
\newblock In {\em Conference on Neural Information Processing Systems
  (NeurIPS)}, 2021.

\bibitem{lei2021clip-bert}
Jie Lei, Linjie Li, Luowei Zhou, Zhe Gan, Tamara~L. Berg, Mohit Bansal, and
  Jingjing Liu.
\newblock {Less is More: ClipBERT for Video-and-Language Learning via Sparse
  Sampling}.
\newblock In {\em Conference on Computer Vision and Pattern Recognition
  (CVPR)}, 2022.

\bibitem{lei2018tvqa}
Jie Lei, Licheng Yu, Mohit Bansal, and Tamara~L. Berg.
\newblock {TVQA: Localized, Compositional Video Question Answering}.
\newblock In {\em Conference on Empirical Methods in Natural Language
  Processing (EMNLP)}, 2018.

\bibitem{lei2020tvqa+}
Jie Lei, Licheng Yu, Tamara~L. Berg, and Mohit Bansal.
\newblock {TVQA+: Spatio-Temporal Grounding for Video Question Answering}.
\newblock In {\em Annual Meeting of the Association for Computational
  Linguistics (ACL)}, 2020.

\bibitem{lei2020tvr}
Jie Lei, Licheng Yu, Tamara~L. Berg, and Mohit Bansal.
\newblock {TVR: A Large-Scale Dataset for Video-Subtitle Moment Retrieval}.
\newblock In {\em European Conference on Computer Vision (ECCV)}, 2020.

\bibitem{li2022alpro}
Dongxu Li, Junnan Li, Hongdong Li, Juan~Carlos Niebles, and Steven~C.H. Hoi.
\newblock {Align and Prompt: Video-and-Language Pre-training with Entity
  Prompts}.
\newblock In {\em Conference on Computer Vision and Pattern Recognition
  (CVPR)}, 2022.

\bibitem{li2020hero}
Linjie Li, Yen-Chun Chen, Yu Cheng, Zhe Gan, Licheng Yu, and Jingjing Liu.
\newblock {HERO: Hierarchical Encoder for Video+Language Omni-representation
  Pre-training}.
\newblock In {\em Conference on Empirical Methods in Natural Language
  Processing (EMNLP)}, 2020.

\bibitem{li2022lavender}
Linjie Li, Zhe Gan, Kevin Lin, Chung-Ching Lin, Zicheng Liu, Ce Liu, and Lijuan
  Wang.
\newblock Lavender: Unifying video-language understanding as masked language
  modeling.
\newblock {\em arXiv preprint arXiv:2206.07160}, 2022.

\bibitem{li2021value}
Linjie Li, Jie Lei, Zhe Gan, Licheng Yu, Yen-Chun Chen, Rohit Pillai, Yu Cheng,
  Luowei Zhou, Xin~Eric Wang, William~Yang Wang, Tamara~Lee Berg, Mohit Bansal,
  Jingjing Liu, Lijuan Wang, and Zicheng Liu.
\newblock {VALUE: A Multi-Task Benchmark for Video-and-Language Understanding
  Evaluation}.
\newblock In {\em Conference on Neural Information Processing Systems
  (NeurIPS)}, 2021.

\bibitem{lin2022swin-bert}
Kevin Lin, Linjie Li, Chung-Ching Lin, Faisal Ahmed, Zhe Gan, Zicheng Liu,
  Yumao Lu, and Lijuan Wang.
\newblock {SwinBERT: End-to-End Transformers with Sparse Attention for Video
  Captioning}.
\newblock In {\em Conference on Computer Vision and Pattern Recognition
  (CVPR)}, 2022.

\bibitem{lin2022frozen}
Ziyi Lin, Shijie Geng, Renrui Zhang, Peng Gao, Gerard de Melo, Xiaogang Wang,
  Jifeng Dai, Yu Qiao, and Hongsheng Li.
\newblock {Frozen Clip Models are Efficient Video Learners}.
\newblock In {\em European Conference on Computer Vision (ECCV)}, 2022.

\bibitem{liu2021hit}
Song Liu, Haoqi Fan, Shengsheng Qian, Yiru Chen, Wenkui Ding, and Zhongyuan
  Wang.
\newblock {HiT: Hierarchical Transformer with Momentum Contrast for Video-Text
  Retrieval}.
\newblock In {\em arXiv:2103.15049}, 2021.

\bibitem{liu2020collaborative-expert}
Yang Liu, Samuel Albanie, Arsha Nagrani, and Andrew Zisserman.
\newblock {Use What You Have: Video Retrieval Using Representations From
  Collaborative Experts}.
\newblock In {\em British Machine Vision Conference (BMVC)}, 2020.

\bibitem{liu2019roberta}
Yinhan Liu, Myle Ott, Naman Goyal, Jingfei Du, Mandar Joshi, Danqi Chen, Omer
  Levy, Mike Lewis, Luke Zettlemoyer, and Veselin Stoyanov.
\newblock {RoBERTa: A Robustly Optimized BERT Pretraining Approach}.
\newblock In {\em arXiv:1907.11692}, 2019.

\bibitem{liu2021swin}
Ze Liu, Yutong Lin, Yue Cao, Han Hu, Yixuan Wei, Zheng Zhang, Stephen Lin, and
  Baining Guo.
\newblock {Swin Transformer: Hierarchical Vision Transformer using Shifted
  Windows}.
\newblock In {\em International Conference on Computer Vision (ICCV)}, 2021.

\bibitem{liu2022video-swin}
Ze Liu, Jia Ning, Yue Cao, Yixuan Wei, Zheng Zhang, Stephen Lin, and Han Hu.
\newblock {Video Swin Transformer}.
\newblock In {\em Conference on Computer Vision and Pattern Recognition
  (CVPR)}, 2022.

\bibitem{loshchilov2019adamw}
Ilya Loshchilov and Frank Hutter.
\newblock {Decoupled Weight Decay Regularization}.
\newblock In {\em International Conference for Learning Representations
  (ICLR)}, 2019.

\bibitem{luo2021clip4clip}
Huaishao Luo, Lei Ji, Ming Zhong, Yang Chen, Wen Lei, Nan Duan, and Tianrui Li.
\newblock {CLIP4Clip: An Empirical Study of CLIP for End to End Video Clip
  Retrieval}.
\newblock In {\em arXiv:2104.08860}, 2021.

\bibitem{miech2020end}
Antoine Miech, Jean-Baptiste Alayrac, Lucas Smaira, Ivan Laptev, Josef Sivic,
  and Andrew Zisserman.
\newblock {End-to-End Learning of Visual Representations from Uncurated
  Instructional Videos}.
\newblock In {\em Conference on Computer Vision and Pattern Recognition
  (CVPR)}, 2020.

\bibitem{miech2019howto100m}
Antoine Miech, Dimitri Zhukov, Jean-Baptiste Alayrac, Makarand Tapaswi, Ivan
  Laptev, and Josef Sivic.
\newblock {HowTo100M: Learning a Text-Video Embedding by Watching Hundred
  Million Narrated Video Clips}.
\newblock In {\em International Conference on Computer Vision (ICCV)}, 2019.

\bibitem{ni2022expanding}
Bolin Ni, Houwen Peng, Minghao Chen, Songyang Zhang, Gaofeng Meng, Jianlong Fu,
  Shiming Xiang, and Haibin Ling.
\newblock {Expanding Language-Image Pretrained Models for General Video
  Recognition}.
\newblock In {\em European Conference on Computer Vision (ECCV)}, 2022.

\bibitem{paszke2019pytorch}
Adam Paszke, Sam Gross, Francisco Massa, Adam Lerer, James Bradbury, Gregory
  Chanan, Trevor Killeen, Zeming Lin, Natalia Gimelshein, Luca Antiga, Alban
  Desmaison, Andreas Köpf, Edward Yang, Zach DeVito, Martin Raison, Alykhan
  Tejani, Sasank Chilamkurthy, Benoit Steiner, Lu Fang, Junjie Bai, and Soumith
  Chintala.
\newblock {PyTorch: An Imperative Style, High-Performance Deep Learning
  Library}.
\newblock In {\em Conference on Neural Information Processing Systems
  (NeurIPS)}, 2019.

\bibitem{patrick2021support-set}
Mandela Patrick, Po-Yao Huang, Yuki Asano, Florian Metze, Alexander Hauptmann,
  Joao Henriques, and Andrea Vedaldi.
\newblock {Support-set bottlenecks for video-text representation learning}.
\newblock In {\em International Conference for Learning Representations
  (ICLR)}, 2021.

\bibitem{radford2021clip}
Alec Radford, Jong~Wook Kim, Chris Hallacy, Aditya Ramesh, Gabriel Goh,
  Sandhini Agarwal, Girish Sastry, Amanda Askell, Pamela Mishkin, Jack Clark,
  Gretchen Krueger, and Ilya Sutskever.
\newblock {Learning Transferable Visual Models From Natural Language
  Supervision}.
\newblock In {\em International Conference on Machine Learning (ICML)}, 2021.

\bibitem{ramesh2021dalle}
Aditya Ramesh, Mikhail Pavlov, Gabriel Goh, Scott Gray, Chelsea Voss, Alec
  Radford, Mark Chen, and Ilya Sutskever.
\newblock {Zero-Shot Text-to-Image Generation}.
\newblock In {\em International Conference on Machine Learning (ICML)}, 2021.

\bibitem{ranftl2021dpt}
Rene Ranftl, Alexey Bochkovskiy, and Vladlen Koltun.
\newblock {Vision Transformers for Dense Prediction}.
\newblock In {\em International Conference on Computer Vision (ICCV)}, 2021.

\bibitem{rasley2020deepspeed}
Jeff Rasley, Samyam Rajbhandari, Olatunji Ruwase, and Yuxiong He.
\newblock {DeepSpeed: System Optimizations Enable Training Deep Learning Models
  with Over 100 Billion Parameters}.
\newblock In {\em Knowledge Discovery in Database (KDD)}, 2020.

\bibitem{rohrbach2015lsmdc}
Anna Rohrbach, Marcus Rohrbach, Niket Tandon, and Bernt Schiele.
\newblock {A Dataset for Movie Description}.
\newblock In {\em Conference on Computer Vision and Pattern Recognition
  (CVPR)}, 2015.

\bibitem{rouditchenko2021avlnet}
Andrew Rouditchenko, Angie Boggust, David Harwath, Brian Chen, Dhiraj Joshi,
  Samuel Thomas, Kartik Audhkhasi, Hilde Kuehne, Rameswar Panda, Rogerio Feris,
  Brian Kingsbury, Michael Picheny, Antonio Torralba, and James Glass.
\newblock {AVLnet: Learning Audio-Visual Language Representations from
  Instructional Videos}.
\newblock In {\em INTERSPEECH}, 2021.

\bibitem{seo2022mv-gpt}
Paul~Hongsuck Seo, Arsha Nagrani, Anurag Arnab, and Cordelia Schmid.
\newblock {End-to-end Generative Pretraining for Multimodal Video Captioning}.
\newblock In {\em Conference on Computer Vision and Pattern Recognition
  (CVPR)}, 2022.

\bibitem{sharma2018cc}
Piyush Sharma, Nan Ding, Sebastian Goodman, and Radu Soricut.
\newblock {Conceptual Captions: A Cleaned, Hypernymed, Image Alt-text Dataset
  For Automatic Image Captioning}.
\newblock In {\em Annual Meeting of the Association for Computational
  Linguistics (ACL)}, 2018.

\bibitem{sun2019videobert}
Chen Sun, Austin Myers, Carl Vondrick, Kevin Murphy, and Cordelia Schmid.
\newblock {VideoBERT: A Joint Model for Video and Language Representation
  Learning}.
\newblock In {\em International Conference on Computer Vision (ICCV)}, 2019.

\bibitem{tan2019lxmert}
Hao Tan and Mohit Bansal.
\newblock {LXMERT: Learning Cross-Modality Encoder Representations from
  Transformers}.
\newblock In {\em Conference on Empirical Methods in Natural Language
  Processing (EMNLP)}, 2019.

\bibitem{tan2021vimpac}
Hao Tan, Jie Lei, Thomas Wolf, and Mohit Bansal.
\newblock {VIMPAC: Video Pre-Training via Masked Token Prediction and
  Contrastive Learning}.
\newblock In {\em arXiv:2106.11250}, 2021.

\bibitem{teed2020raft}
Zachary Teed and Jia Deng.
\newblock {RAFT: Recurrent All-Pairs Field Transforms for Optical Flow}.
\newblock In {\em European Conference on Computer Vision (ECCV)}, 2020.

\bibitem{tong22video-mae}
Zhan Tong, Yibing Song, Jue Wang, and Limin Wang.
\newblock {VideoMAE: Masked Autoencoders are Data-Efficient Learners for
  Self-Supervised Video Pre-Training}.
\newblock In {\em arXiv:2203.12602}, 2022.

\bibitem{torabi2016lsmdc-fib}
Atousa Torabi, Niket Tandon, and Leonid Sigal.
\newblock {Learning Language-Visual Embedding for Movie Understanding with
  Natural-Language}.
\newblock In {\em arXiv:1609.08124}, 2016.

\bibitem{touvron2021training}
Hugo Touvron, Matthieu Cord, Matthijs Douze, Francisco Massa, Alexandre
  Sablayrolles, and Herve Jegou.
\newblock {Training Data-efficient Image Transformers \& Distillation through
  Attention}.
\newblock In {\em International Conference on Machine Learning (ICML)}, 2021.

\bibitem{oord2017vq-vae}
Aaron van~den Oord, Oriol Vinyals, and Koray Kavukcuoglu.
\newblock {Neural Discrete Representation Learning}.
\newblock In {\em Conference on Neural Information Processing Systems
  (NeurIPS)}, 2017.

\bibitem{vaswani2017attention}
Ashish Vaswani, Noam Shazeer, Niki Parmar, Jakob Uszkoreit, Llion Jones,
  Aidan~N Gomez, {\L}ukasz Kaiser, and Illia Polosukhin.
\newblock {Attention Is All You Need}.
\newblock In {\em Conference on Neural Information Processing Systems
  (NeurIPS)}, 2017.

\bibitem{wang2022all-in-one}
Alex~Jinpeng Wang, Yixiao Ge, Rui Yan, Yuying Ge, Xudong Lin, Guanyu Cai,
  Jianping Wu, Ying Shan, Xiaohu Qie, and Mike~Zheng Shou.
\newblock {All in One: Exploring Unified Video-Language Pre-training}.
\newblock In {\em arXiv:2203.07303}, 2022.

\bibitem{wang2016temporal}
Limin Wang, Yuanjun Xiong, Zhe Wang, Yu Qiao, Dahua Lin, Xiaoou Tang, and
  Luc~Van Gool.
\newblock {Temporal Segment Networks: Towards Good Practices for Deep Action
  Recognition}.
\newblock In {\em European Conference on Computer Vision (ECCV)}, 2016.

\bibitem{wang2022bevt}
Rui Wang, Dongdong Chen, Zuxuan Wu, Yinpeng Chen, Xiyang Dai, Mengchen Liu,
  Yu-Gang Jiang, Luowei Zhou, and Lu Yuan.
\newblock {BEVT: BERT Pretraining of Video Transformers}.
\newblock In {\em Conference on Computer Vision and Pattern Recognition
  (CVPR)}, 2022.

\bibitem{wang2019vatex}
Xin Wang, Jiawei Wu, Junkun Chen, Lei Li, Yuan-Fang Wang, and William~Yang
  Wang.
\newblock {VATEX: A Large-Scale, High-Quality Multilingual Dataset for
  Video-and-Language Research}.
\newblock In {\em International Conference on Computer Vision (ICCV)}, 2019.

\bibitem{wei2021masked-feat}
Chen Wei, Haoqi Fan, Saining Xie, Chao-Yuan Wu, Alan Yuille, and andChristoph
  Feichtenhofer.
\newblock {Masked Feature Prediction for Self-Supervised Visual Pre-Training}.
\newblock In {\em arXiv:2112.09133}, 2022.

\bibitem{wei2022mvp}
Longhui Wei, Lingxi Xie, Wengang Zhou, Houqiang Li, and Qi Tian.
\newblock {MVP: Multimodality-guided Visual Pre-training}.
\newblock In {\em arXiv:2203.05175}, 2022.

\bibitem{xie2018rethinking}
Saining Xie, Chen Sun, Jonathan Huang, Zhuowen Tu, and Kevin Murphy.
\newblock {Rethinking Spatiotemporal Feature Learning: Speed-Accuracy
  Trade-offs in Video Classification}.
\newblock In {\em European Conference on Computer Vision (ECCV)}, 2018.

\bibitem{xie2022simmim}
Zhenda Xie, Zheng Zhang, Yue Cao, Yutong Lin, Jianmin Bao, Zhuliang Yao, Qi
  Dai, and Han Hu.
\newblock {SimMIM: A Simple Framework for Masked Image Modeling}.
\newblock In {\em Conference on Computer Vision and Pattern Recognition
  (CVPR)}, 2022.

\bibitem{xu2017msrvtt-qa}
Dejing Xu, Zhou Zhao, Jun Xiao, Fei Wu, Hanwang Zhang, Xiangnan He, and Yueting
  Zhuang.
\newblock {Video Question Answering via Gradually Refined Attention over
  Appearance and Motion}.
\newblock In {\em ACM Multimedia (ACMMM)}, 2017.

\bibitem{xu2016msrvtt}
Jun Xu, Tao Mei, Ting Yao, and Yong Rui.
\newblock {MSR-VTT: A Large Video Description Dataset for Bridging Video and
  Language}.
\newblock In {\em Conference on Computer Vision and Pattern Recognition
  (CVPR)}, 2016.

\bibitem{yang2021just-ask}
Antoine Yang, Antoine Miech, Josef Sivic, Ivan Laptev, and Cordelia Schmid.
\newblock {Just Ask: Learning to Answer Questions from Millions of Narrated
  Videos}.
\newblock In {\em International Conference on Computer Vision (ICCV)}, 2021.

\bibitem{yang2020bert-vqa}
Zekun Yang, Noa Garcia, Chenhui Chu, Mayu Otani, Yuta Nakashima, and Haruo
  Takemura.
\newblock {BERT Representations for Video Question Answering}.
\newblock In {\em Winter Conference on Applications of Computer Vision (WACV)},
  2020.

\bibitem{yu2018js-fusion}
Youngjae Yu, Jongseok Kim, and Gunhee Kim.
\newblock {A Joint Sequence Fusion Model for Video Question Answering and
  Retrieval}.
\newblock In {\em European Conference on Computer Vision (ECCV)}, 2018.

\bibitem{zellers2021merlot}
Rowan Zellers, Ximing Lu, Jack Hessel, Youngjae Yu, Jae~Sung Park, Jize Cao,
  Ali Farhadi, and Yejin Choi.
\newblock {MERLOT: Multimodal Neural Script Knowledge Models}.
\newblock In {\em Conference on Neural Information Processing Systems
  (NeurIPS)}, 2021.

\bibitem{zhang2018video-text}
Bowen Zhang, Hexiang Hu, and Fei Sha.
\newblock {Cross-Modal and Hierarchical Modeling of Video and Text}.
\newblock In {\em European Conference on Computer Vision (ECCV)}, 2018.

\bibitem{zhou2022ibot}
Jinghao Zhou, Chen Wei, Huiyu Wang, Wei Shen, Cihang Xie, Alan Yuille, and Tao
  Kong.
\newblock {iBOT: Image BERT Pre-Training with Online Tokenizer}.
\newblock In {\em International Conference on Learning Representations (ICLR)},
  2012.

\bibitem{zhou2018youcook2}
Luowei Zhou, Chenliang Xu, and Jason~J. Corso.
\newblock {Towards Automatic Learning of Procedures from Web Instructional
  Videos}.
\newblock In {\em AAAI Conference on Artificial Intelligence (AAAI)}, 2018.

\bibitem{zhu2020act-bert}
Linchao Zhu and Yi Yang.
\newblock {ActBERT: Learning Global-Local Video-Text Representations}.
\newblock In {\em Conference on Computer Vision and Pattern Recognition
  (CVPR)}, 2020.

\end{thebibliography}
